\documentclass{article}

\PassOptionsToPackage{numbers}{natbib}
\usepackage[preprint]{neurips_2025}

\usepackage[utf8]{inputenc} %
\usepackage[T1]{fontenc}    %
\usepackage{hyperref}       %
\usepackage{url}            %
\usepackage{booktabs}       %
\usepackage{amsfonts}       %
\usepackage{nicefrac}       %
\usepackage{microtype}      %
\usepackage{xcolor}         %
\usepackage{enumitem}
\usepackage{booktabs}
\usepackage{caption}
\usepackage{float}
\usepackage{multirow}
\usepackage{graphicx}
\usepackage{amsthm}
\usepackage{wrapfig}
\usepackage{lipsum}
\usepackage{amsmath}
\usepackage{wrapfig}
\usepackage{makecell}
\usepackage{pifont}
\usepackage{stmaryrd} %

\usepackage{algorithm}
\usepackage{algpseudocode}
\usepackage{xr}
\usepackage{bbm}
\usepackage{subcaption}

\newtheorem{definition}{Definition}

\newcommand{\modelname}{\texttt{MiCADangelo}}

\makeatletter
\def\maketitlesupplementary{%
  \newpage
  \vbox{%
    \hsize\textwidth
    \linewidth\hsize
    \vskip 0.1in
    \@toptitlebar
    \centering
    {\LARGE\bf \@title\par}
    \vspace{0.5em}
    {\Large Supplementary Material\par}
    \@bottomtitlebar
  }
}
\makeatother

\title{\texttt{MiCADangelo}: Fine-Grained Reconstruction of Constrained CAD Models from 3D Scans}

\author{%
  Ahmet Serdar Karadeniz \\
  SnT, University of Luxembourg\\
  \texttt{ahmet.karadeniz@uni.lu} \\
  \And
  Dimitrios Mallis \\
  SnT, University of Luxembourg\\
  \texttt{dimitrios.mallis@uni.lu} \\
  \And
  Danila Rukhovich \\
  SnT, University of Luxembourg\\
  \texttt{danila.rukhovich@uni.lu}
  \And
  Kseniya Cherenkova \\
  SnT, University of Luxembourg, Artec 3D\\
  \texttt{kseniya.cherenkova@uni.lu} \\
  \And
  Anis Kacem \\
  SnT, University of Luxembourg\\
  \texttt{anis.kacem@uni.lu}
  \And
  Djamila Aouada\\
  SnT, University of Luxembourg\\
  \texttt{djamila.aouada@uni.lu}
}

\begin{document}

\maketitle
\begin{abstract}

Computer-Aided Design (CAD) plays a foundational role in modern manufacturing and product development, often requiring designers to modify or build upon existing models. Converting 3D scans into parametric CAD representations—a process known as CAD reverse engineering—remains a significant challenge due to the high precision and structural complexity of CAD models. Existing deep learning-based approaches typically fall into two categories: bottom-up, geometry-driven methods, which often fail to produce fully parametric outputs, and top-down strategies, which tend to overlook fine-grained geometric details. Moreover, current methods neglect an essential aspect of CAD modeling: sketch-level constraints. In this work, we introduce a novel approach to CAD reverse engineering inspired by how human designers manually perform the task. Our method leverages multi-plane cross-sections to extract 2D patterns and capture fine parametric details more effectively. It enables the reconstruction of detailed and editable CAD models, outperforming state-of-the-art methods and, for the first time, incorporating sketch constraints directly into the reconstruction process.
\end{abstract}
    
\section{Introduction}
\label{sec:intro}

Computer-Aided Design (CAD) modeling, typically performed using specialized software~\cite{Solidworks, Onshape, FreeCAD}, plays a critical role in the development and manufacturing of real-world objects. In modern CAD workflows, designers begin by creating 2D sketches composed of parametric curves (e.g., lines, arcs, circles), which are further constrained by geometric relationships (e.g., coincident, tangent, parallel)~\cite{seff2022vitruvion}. These sketches are then transformed into 3D geometry using operations such as extrusion, revolution, and cutting. By following this sequential process, designers can construct 3D parametric solids that accurately represent the intended object. A common and practical approach in CAD workflows is to begin with a template of an existing object and adapt it to new design requirements. However, existing real-world objects commonly do not have publicly available CAD models. In such cases, a digital representation can be obtained through 3D scanning, which usually produces a 3D mesh. While meshes capture the surface geometry of an object, they are unstructured and lack the parametric information needed for further design and modification in CAD software. Consequently, converting a 3D mesh into a parametric CAD model—an operation commonly referred to as \textit{CAD reverse engineering}—is a crucial step for enabling efficient design iteration and customization.%

\begin{figure}[t]
    \centering
\setlength{\belowcaptionskip}{-0.5cm}
    \includegraphics[width=\linewidth]{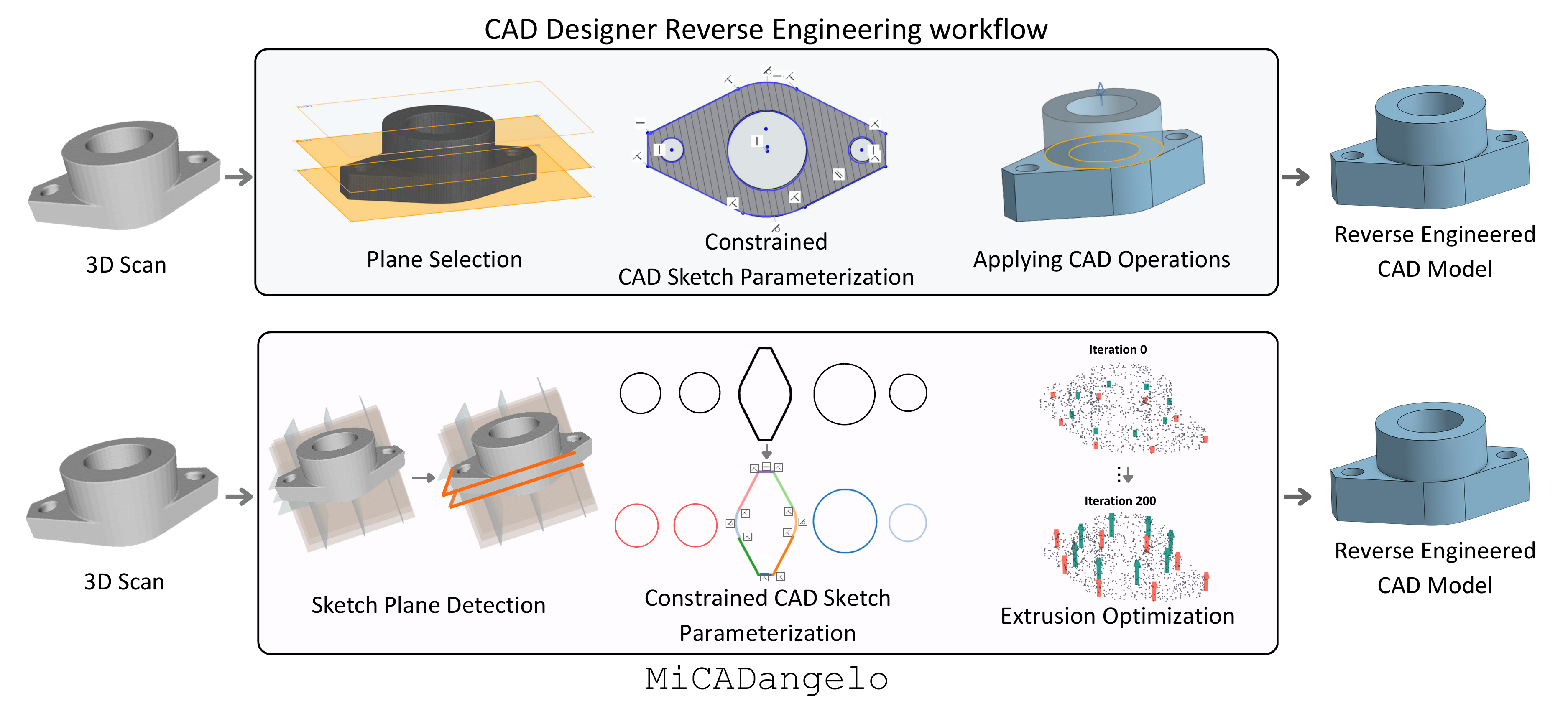}
    \caption{\modelname~is a novel framework for CAD reverse-engineering that mimics human design workflows. It analyzes 3D scans via 2D cross-sections to detect sketch planes, predict constrained parametric sketches and optimize extrusions.%
    }
    
    \label{fig:teaser}
\end{figure}

Today, CAD reverse engineering is typically carried out manually by designers using specialized software~\cite{polyga2025}. As illustrated in the top part of Figure~\ref{fig:teaser}, the process generally involves the following steps: (i) extracting a 2D cross-section of a specific part from the imported 3D scan; (ii) reconstructing the contour of the section using parametric 2D curves and placing the CAD constraints to restrict the modification; and (iii) applying CAD operations (e.g., extrusion) to the parametric sketch to create a 3D solid model of the part. These steps are repeated for the remaining parts until the entire 3D scan is reconstructed as a complete parametric CAD model within the software. While this process allows for faster CAD modeling compared to creating models from scratch, it remains tedious and demands CAD expertise. As a result, automating this process became an active research area~\cite{wu2021deepcad,ren2022extrudenet,uy2022point2cyl,ma2024draw,lambourne2022reconstructing, rukhovich2025cad, mallis2025cad}. 

Automated 3D CAD reverse engineering has advanced from identifying parts in 3D scans~\cite{li2019supervised, sharma2020parsenet}, to fitting parametric primitives to scanned geometry~\cite{wang2020pie, cherenkova2023sepicnet}, and more recently, to predicting sequential design step directly from point clouds~\cite{khan2024cad, ma2024draw}. The latter approach is particularly valuable, as it supports parametric reconstruction while preserving an editable design history. This advancement has been enabled by the emergence of datasets such as DeepCAD~\cite{wu2021deepcad} and Fusion 360~\cite{willis2020fusion}, which provide sequences of 2D sketches and extrusion operations—commonly referred to as \textit{sketch-extrude} sequences. Several recent methods have explored learning the mapping from point clouds to sketch-extrude sequences, which can be broadly categorized into: (i) top-down approaches~\cite{khan2024cad, ma2024draw, xu2023hierarchical}, which directly predict the parameters of the sketch-extrude sequence as a series of parametric values from the input point cloud; and (ii) bottom-up approaches~\cite{uy2022point2cyl, li2023secad, ren2022extrudenet}, which attempt to predict individual extrusion cylinders that, when combined, closely approximate the target geometry. While top-down approaches offer the advantage of producing fully parametric outputs that approximate ground-truth sketch-extrude sequences, they often struggle to capture fine-grained details of the input geometry frequently dominated by larger structures. In contrast, bottom-up approaches can better preserve local geometrical details, but they lack full parametrization and do not integrate seamlessly into standard CAD workflows~\cite{khan2024cad}. %
Furthermore, to the best of our knowledge, none of these approaches consider CAD sketch constraints which are a critical component of CAD modeling~\cite{seff2022vitruvion,ganin2021computer}.

In this work, we propose \modelname, a solution for automating CAD reverse engineering by emulating the way human designers approach the task. As illustrated in the bottom part of Figure.\ref{fig:teaser}, \modelname~begins by analyzing the input scan through a series of 2D cross-sectional slices and predicting key slices to serve as sketch planes. For each selected plane, closed loops are extracted from the cross-section and converted into raster images, which are used to predict both the 2D parametric curves and associated CAD sketch constraints. These constrained sketches are then extruded via an optimization process that leverages the local mesh geometry around each loop. The resulting extruded parts are ultimately merged to produce a structured, fully parametric CAD model. %
\modelname~offers the advantage of preserving fine-grained geometric details of the input scan while generating fully parametric sketch-extrude sequences, including CAD constraints—an aspect largely unexplored in 3D CAD reverse engineering.

\textbf{Contributions} The main contributions of this work can be summarized as follows:
\vspace{-0.35cm}
\begin{enumerate}[itemsep=2pt, parsep=0pt]
    \item We introduce a real-world CAD reverse engineering-inspired approach that can effectively reconstruct fully parametric CAD models from 3D scans while preserving fine-grained geometric details.
    \item To the best of our knowledge, this is the first approach capable of reconstructing CAD models from 3D scans while incorporating sketch constraints.
    \item We conduct comprehensive experiments on publicly available benchmarks and demonstrate that our method outperforms existing state-of-the-art techniques in CAD reverse engineering.

\end{enumerate}

\section{Related Work}
\label{sec:related_work}

\textbf{3D CAD Reverse Engineering.} Early approaches to CAD reconstruction rely on Constructive Solid Geometry (CSG) representations~\cite{du2018inversecsg, friedrich2019optimizing, kania2020ucsg, geometry2022neural, yu2022capri, yu2023d, ren2021csg}, using Boolean operations over primitives, but these are limited in capturing complex, real-world CAD structures. Other works target Boundary Representation (BRep) reconstruction~\cite{lambourne2021brepnet, xu2021inferring, willis2020fusion, jayaraman2021uv}, focusing on high-precision surface and topology modeling. More recent efforts adopt sketch-extrude paradigms, which better reflect the way CAD models are constructed in practice. Within this direction, geometry-grounded bottom-up methods~\cite{li2023secad, ren2022extrudenet, uy2022point2cyl} estimate sketches and extrusion parameters from input scans, leveraging structural cues from the geometry but often lacking full parametric editability and seamless integration in CAD software~\cite{khan2024cad}. Alternatively, language-based top-down methods~\cite{khan2024cad, xu2023hierarchical, ma2024draw} learn to generate construction sequences, capturing design intent but struggling to preserve fine-grained geometric fidelity. We propose a practical reverse engineering-inspired approach that reconstructs detailed and editable CAD models with explicit sketch primitives and constraints, directly compatible with standard CAD software.

\textbf{Constrained Sketches in CAD.} In CAD modeling, sketch primitives and their associated constraints form the foundation of design intent~\cite{Zhang2009DesignII, Otey2018RevisitingTD}, as they govern how a design adapts when modified. Parametric CAD sketch primitives (e.g., lines, arcs, circles) provide a flexible 2D interface for defining shapes, while constraints (e.g., perpendicular, parallel, tangent) preserve geometric relationships, allowing controlled variation without compromising the overall structure. When designers apply constraints (e.g., orthogonality) to sketch primitives (e.g., two lines), any changes made to one primitive are automatically adjusted to preserve these relationships. For example, if the parameters of one line are modified, the other will update accordingly to maintain orthogonality. In a reverse engineering context, it is highly desirable not only to infer accurate CAD geometry from input scans but also to recover the underlying design intent, ensuring that the reconstructed model responds to modifications in a manner consistent with how a designer would have originally constrained it. The availability of large-scale CAD sketch datasets~\cite{seff2020sketchgraphs} has facilitated research in this area, with recent work focusing on generative CAD sketch modeling~\cite{ganin2021computer, seff2022vitruvion, para2021sketchgen}, sketch image parameterization~\cite{karadeniz2025picasso, karadeniz2024davinci, wu2024cadvlm}, and design intent inference~\cite{yang2022discovering, casey2025aligning}. While these methods have achieved promising results for 2D sketches, their effectiveness in 3D CAD modeling remains underexplored.

\section{Preliminaries and Problem Statement}
\label{sec:problem_statement}

In this section, we provide the problem statement along with the preliminary definitions. %

\begin{definition}[Sketch Primitive]
A sketch primitive $\mathbf{p}$ is a 2D entity uniquely defined by geometric parameters. We define the following 2D sketch primitives: \textbf{Line:} Defined by its start point $(x_s, y_s)~\in~\mathbb{R}^2$ and end point $(x_e, y_e)~\in~\mathbb{R}^2$. \textbf{Circle:} Represented by its center $(x_c, y_c) \in \mathbb{R}^2$ and radius $r \in \mathbb{R}$. \textbf{Arc:} A segment of a circle, specified by its start point $(x_s, y_s) \in \mathbb{R}^2$, mid-point $(x_m, y_m) \in \mathbb{R}^2$, and end point $(x_e, y_e) \in \mathbb{R}^2$, all lying on the same circle.
\end{definition}

\begin{definition}[Sketch Constraint]
A sketch constraint $\mathbf{c} = (\mathbf{p}_i,\mathbf{p}_j, c_t)$ is defined by a constraint type $c_t$ that specifies the relationship between two primitives $\mathbf{p}_i$ and $\mathbf{p}_j$. The types of constraints considered in this work are: coincident, concentric, equal, fix, horizontal, midpoint, normal, offset, parallel, perpendicular, quadrant, tangent, and vertical. Note that some constraints are defined on a single primitive such as vertical and horizontal. 

\end{definition}

\begin{definition}[Constrained Sketch]
A constrained sketch $\mathcal{K} = (\mathcal{P}, \mathcal{C})$ is defined by a set of $n_p$ 2D sketch primitives $\mathcal{P} = \{\mathbf{p}_i\}_{i=1}^{n_p}$ and a set of $n_c$ sketch constraints $\mathcal{C} = \{\mathbf{c}_i\}_{i=1}^{n_c}$.
\end{definition}

\begin{definition}[Sketch Plane]
A sketch plane $\pi$ is defined by an origin point $\mathbf{o} \in \mathbb{R}^3$ and a normal vector $\mathbf{n} \in \mathbb{R}^3$, on which a sketch is defined and centered at $\mathbf{o}$.
\end{definition}

\begin{definition}[3D Mesh]
 A 3D mesh is defined as $\mathbf{M}= (\mathcal{V}, \mathcal{F})$ where $\mathcal{V} \subset \mathbb{R}^3$ are vertices and $\mathcal{F} \subset \mathcal{V}^3$ are triangles formed by 3 vertices. 
 \end{definition}

\begin{definition}[Cross-section Slice]

Given a 3D mesh $\mathbf{M}$ and a slicing plane $\pi$, a cross-section slice $\mathcal{S}$ is formed by sets of line segments resulting from the intersection of the mesh triangles with $\pi$. This slicing process naturally yields multiple connected components, each corresponding to a distinct set of connected line segments. Formally, the cross-section slice $\mathcal{S}$ can be represented as a collection of $L$ line sets $\{\mathbb{L}_j\}_{j=1}^{L}$, where each line set composed of $n_l$ successive points $ \{\mathbf{q}_i\}_{i=1}^{n_l} \in \mathbb{R}^3$ is defined as  $\mathbb{L}_j~=~\left\{ (\mathbf{q}_i, \mathbf{q}_{i+1}) \mid i = 1, \dots, n_l - 1 \right\}$.

\end{definition}

\begin{definition}[Closed Loop]
A closed loop $\mathbf{L}$ is a set of line segments connecting $n_l$ successive points $ \{\mathbf{q}_i\}_{i=1}^{n_l} \in \mathbb{R}^3$ forming a non-self-intersecting and enclosed region. It is defined as: $
\mathbf{L}~=~\left\{ (\mathbf{q}_i, \mathbf{q}_{i+1}) \mid i = 1, \dots, n_l-1 \right\} \cup \left\{ (\mathbf{q}_{n_l}, \mathbf{q}_1) \right\}$. It is important to note that, in the context of CAD models, the line sets forming the cross-section slice $\mathcal{S}$ defined in the above definition are usually closed loops.

\end{definition}

\begin{definition}[Extrusion]
An extrusion $\mathbf{e} = (\pi, t, \mathbf{v}, h)$ is defined by a sketch plane $\pi$, an extrusion type $t$, and extrusion parameters $(\mathbf{v}, h)$, where $\mathbf{v} \in \mathbb{R}^3$ is the extrusion direction vector and $h \in \mathbb{R}$ is the extrusion length. The extrusion operation extends a sketch $\mathcal{K}$ along $\mathbf{v}$ for a distance $h$, forming a 3D solid. The extrusion type $t$ specifies whether the operation creates material (new) or removes material (cut).
\label{def:extrusion}
\end{definition}

\begin{definition}[CAD Model Representation]
A CAD model $\mathbf{C}$ is defined as the sequential combination of $n_s$ sketch-extrude steps $\{(\mathcal{K}_i,\mathbf{e}_i)\}_{i=1}^{n_s}$.

\end{definition}

\textbf{Problem Statement.} Given an input 3D mesh \( \mathbf{M} \), the goal is to recover the sequence of $n_s$ sketch-extrude steps reconstructing the CAD model \( \mathbf{C} = \{(\mathcal{K}_j,\mathbf{e}_j) \}_{j=1}^{n_s} \). Each extrusion \(\mathbf{e}_j~=~(\pi_j, t_j, \mathbf{v}_j, h_j) \) is defined by a sketch plane $\pi_j$, an extrusion type $t_j$, and extrusion parameters $(\mathbf{v}_j, h_j)$. Each constrained sketch $\mathcal{K}_j = (\mathcal{P}_j,\mathcal{C}_j)$ consists of 2D parametric primitives $\mathcal{P}_j$ along with their associated CAD constraints $\mathcal{C}_j$.

\section{Proposed Method}
\label{sec:proposed_method}

\begin{figure}[t]
    \centering
    \includegraphics[width=\linewidth]{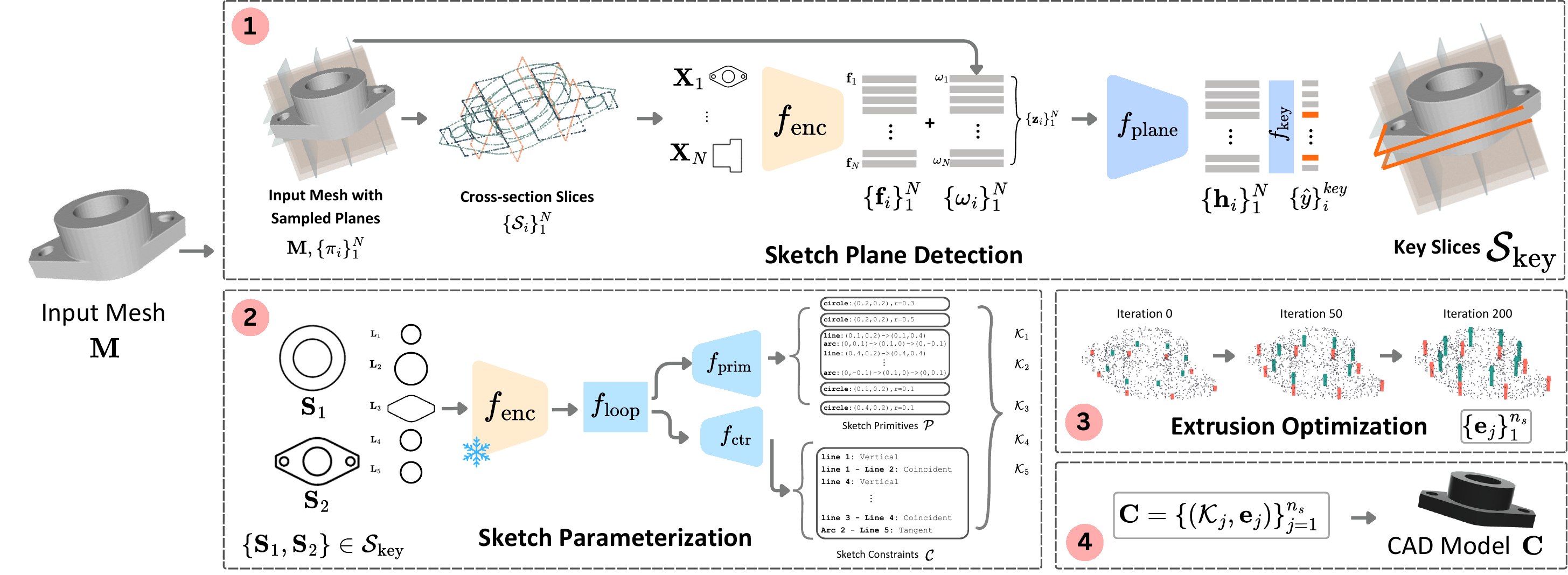}
    \caption{Overview of the method. \modelname \space comprises three main components: Sketch Plane Detection, Sketch Parameterization, and Differentiable Extrusion. The generated constrained sketches, together with the optimized extrusion parameters, are assembled into the final parametric CAD model.}
    \label{fig:method}
\end{figure}

\subsection{Method Overview}

As illustrated in Figure~\ref{fig:teaser}, \modelname~is inspired by the way human designers approach CAD reverse engineering. Given an input mesh \( \mathbf{M} \) obtained from a 3D scan, the key idea—mirroring human intuition—is to identify the most relevant cross-section slices that enable accurate reconstruction of the original CAD model. To achieve this, \modelname~samples $N$ equally-spaced slicing planes \( \{ \pi_i  \}_{i=1}^{N}\) along the \( x \)-, \( y \)-, and \( z \)-axes, producing a set of 2D cross-section slices \( \{ \mathcal{S}_i \}_{i=1}^{N} \). These slices are then converted into raster images and supplied with contextual embeddings before being passed into a sketch plane detection network that identifies the most relevant slices \( \mathcal{S}_{\text{key}} = \{\mathcal{S}_i\}_{i=1}^{n_{key}} \) (Section~\ref{sec:planedet}). Once the key cross-section slices are identified—again following the logic used by human designers—the next step is to predict the sketch primitives and their associated constraints from these slices to obtain the corresponding constrained sketches. To achieve this, each key cross-section slice is decomposed into separate closed loops $\{\mathbf{L}_j\}_{j=1}^L$. Each loop is converted into a raster image and passed to a network for constrained sketch parameterization with the goal of inferring the corresponding sketch primitives and constraints $\{\mathcal{K}_j\}_{j=1}^L$ (Section~\ref{sec:param}). After obtaining the constrained sketches, a differentiable extrusion optimization is performed for each sketch $\mathcal{K}_j$ \textit{w.r.t} the input geometry of the mesh $\mathbf{M}$ to find out the corresponding extrusion parameters $\mathbf{e}_j$ (Section~\ref{sec:extrusion}). Finally, the obtained extruded elements are assembled together to obtain the final CAD model. An overview of the proposed method is illustrated in Figure~\ref{fig:method}.

\vspace{-0.1cm}
\subsection{Sketch Plane Detection Network}
\label{sec:planedet}

Given a set of cross-section slices \( \{\mathcal{S}_i\}_{i=1}^N \) extracted from the input mesh \( \mathbf{M} \), our goal is to identify a subset of these as key sketch plane slices. Each slice \( \mathcal{S}_i \) is projected onto a 2D plane and normalized to fit within a unit bounding box, which is subsequently rendered as a binary image \( \mathbf{X}_i \in \{0,1\}^{H \times W} \). The slice image $\mathbf{X}_i$ is passed through a convolutional ResNet34 encoder $f_{\text{enc}}~:~\{0,1\}^{H \times W} \rightarrow \mathbb{R}^d$ to produce a $d$-dimensional latent vector $\mathbf{f}_i = f_{\text{enc}}(\mathbf{X}_i)$. In order to contextualize the embedding of the different slices and maintain their spatial relationships, each slice is associated with a slice index $\sigma_i \in \{0, \dots, N-1\}$, axis identifier $a_i \in \{0, 1, 2\}$ for $x$-, $y$-, or $z$-axis, normalization parameters $\boldsymbol{\eta}_i = (t^x_i, t^y_i, s_i) \in \mathbb{R}^3$ that correspond to the translation and the scale parameters of the normalization process. These inputs are embedded into the latent space $\mathbb{R}^d$,

\begin{equation}
\mathbf{\omega}_i^{\text{pos}} = \mathbf{W}_{\text{pos}} \mathbf{y}_{\sigma_i}, \quad
\mathbf{\omega}_i^{\text{axis}} = \mathbf{W}_{\text{axis}} \mathbf{y}_{a_i}, \quad
\mathbf{\omega}_i^{\text{norm}} = \mathbf{W}_{\text{norm}} \boldsymbol{\eta}_i \ ,
\end{equation}
where $\mathbf{y}_{\sigma_i} \in \{0,1\}^N$ and $\mathbf{y}_{a_i} \in \{0,1\}^3$ are one-hot vectors for slice index and axis identifier, $\mathbf{W}_{\text{pos}} \in \mathbb{R}^{d \times N}$ and $\mathbf{W}_{\text{axis}} \in \mathbb{R}^{d \times 3}$ are the corresponding learnable matrices, and $\mathbf{W}_{\text{norm}} \in \mathbb{R}^{d \times 3}$ is a linear projection of normalization parameters. The final embedding $ \mathbf{z}_i \in \mathbb{R}^d$ is obtained as
\begin{equation}
    \mathbf{z}_i = \mathbf{f}_i + \mathbf{\omega}_i  \ ,
\end{equation}
where the contextual embedding is defined as 
$\mathbf{\omega}_i = \mathbf{\omega}_i^{\text{pos}} + \mathbf{\omega}_i^{\text{axis}} + \mathbf{\omega}_i^{\text{norm}}$. Next, the set of embedding $\{\mathbf{z}_i\}_{i=1}^N$ is fed into a multi-layer multi-head transformer encoder $f_{\text{plane}}: \mathbb{R}^d \rightarrow \mathbb{R}^d$ to obtain a set of richer embedding $\{\mathbf{h}_i\}_{i=1}^N \in \mathbb{R}^d$ as follows, 
\begin{equation}
    (\mathbf{h}_1, \dots, \mathbf{h}_N) = f_{\text{plane}}(\mathbf{z}_1, \dots, \mathbf{z}_N) \ .
\end{equation}

Each encoded slice embedding $\mathbf{h}_i$ is passed through a binary classifier $f_{\text{key}}: \mathbb{R}^d \rightarrow [0,1]$ composed of a linear layer and a sigmoid to predict the probability of being a key sketch plane as follows,
\begin{equation}
    \hat{y}^{\text{key}}_i = f_{\text{key}}(\mathbf{h}_i) \ . 
\end{equation}

Note that the sketch plane detection network is trained by a binary cross-entropy loss using supervised key sketch plane labels from the DeepCAD dataset~\cite{wu2021deepcad}. The process of obtaining these labels is explained in the supplementary. Finally, the subset \( \mathcal{S}_{\text{key}} = \{\mathcal{S}_i \mid  \hat{y}^{\text{key}}_i \geq \tau \} \) is determined to be the set of key slices with the corresponding planes $\pi_i$, where \( \tau \) is a fixed threshold set to 0.5.

\vspace{-0.1cm}
\subsection{Constrained Sketch Parameterization Network}
\label{sec:param}

Once the key cross-section slices $\mathcal{S}_{\text{key}}$ are obtained by the sketch plane detection network, the goal of the constrained sketch parameterization network is to infer the constrained parametric sketches $\{\mathcal{K}_j\}_{j=1}^L$ from the individual closed loops $\{\mathbf{L}_j\}_{j=1}^L$ of each key cross-section slice. For each $\mathcal{S}_i \in \mathcal{S}_{\text{key}}$, the set of resulting closed loops $\{\mathbf{L}_j\}_{j=1}^L$ are projected to 2D and rendered as binary images $\{\mathbf{X}_{j}\}_{j=1}^L$. The constrained sketch parameterization network operates directly on these rasterized loop images to infer the corresponding sketch primitives and geometric constraints. This approach enables to preserve fine-grained geometric details of the input mesh, which are essential for accurate CAD reconstruction.

The design of the constrained sketch parameterization network is similar to the one introduced in~\cite{karadeniz2024davinci}. Each image $\mathbf{X}_{j} \in \{0,1\}^{H \times W}$ is then encoded by a convolutional encoder $f_{\text{enc}}~:~\{0,1\}^{H \times W} \rightarrow \mathbb{R}^d$ (same as the one used to encode the cross-section slices in Section~\ref{sec:planedet}), producing a latent feature vector $\mathbf{f}_{j} \in \mathbb{R}^d$. The latent vector is then passed to a transformer encoder-decoder $f_{\text{loop}}~:~\mathbb{R}^d \rightarrow \mathbb{R}^{d_e \times n_p }$ to obtain a set of embedding for $n_p$ sketch primitives $\zeta_j~\in \mathbb{R}^{d_e \times n_p }$ as follows, 
\begin{equation}
    \zeta_j = f_{\text{loop}}(\mathbf{f}_j) \ .
\end{equation}

As in~\cite{karadeniz2024davinci}, the set of embedding is passed to two separate heads: a parameterization head $f_{\text{prim}}~:~\mathbb{R}^{d_e \times n_p }\rightarrow \mathbb{Q}^{n_p}$ and a constraint prediction head $f_{\text{ctr}}~:~\mathbb{R}^{d_e \times n_p }\rightarrow \mathbb{Q}^{n_c}$, where $\mathbb{Q}^{n_p}$ and $ \mathbb{Q}^{n_c}$ denote the quantized space of $n_p$ primitive values and  $n_c$ constraint values, respectively. These heads result in the parametric sketch primitives and their constraints as follows,

\begin{equation}
\mathcal{P}_j = f_{\text{prim}}(\zeta_j)   \quad , \quad
\mathcal{C}_j = f_{\text{ctr}}(\zeta_j) \ .
\end{equation}

For further details on the quantization spaces of primitives and constraints, the design of the constrained sketch parameterization network, and the loss functions employed during training, readers are referred to~\cite{karadeniz2024davinci} as well as the supplementary materials for additional explanations. Due to the lack of CAD constraint annotations in the DeepCAD dataset~\cite{wu2021deepcad} and other 3D CAD modeling datasets, the constrained sketch parameterization network is initially trained on the SketchGraphs dataset~\cite{seff2020sketchgraphs}. It is then fine-tuned on an augmented version of SketchGraphs that includes added noise and synthetically generated closed-loop images, enabling better adaptation to the characteristics of cross-sectional slices. Note that both the sketch parameterization network and the sketch plane detection network share the same encoder \( f_{\text{enc}} \). The encoder is initially trained as part of the sketch parameterization network and then kept frozen during parameterization, while it is fine-tuned during the training of the sketch plane detection network.

\vspace{-0.1cm}
\subsection{Differentiable Extrusion Optimization}
\label{sec:extrusion}
For each constrained sketch \( \mathcal{K}_{j} \) corresponding to a loop \( \mathbf{L}_{j} \) within a key cross-section slice $\mathcal{S}_i$, the goal of differentiable extrusion optimization is to recover the parameters of the extrusion $\mathbf{e}_j$ that best fits the resulting extruded solid to the 3D input mesh. As mentioned in Definition~\ref{def:extrusion} of Section~\ref{sec:problem_statement}, the extrusion is defined by $\mathbf{e}_j = (\pi_j, t_j, \mathbf{v}_j, h_j)$, with $\pi_j$ denotes the sketch plane from which the extrusion originates, $t_j \in \{\textit{new}, \textit{cut}\}$ specifies the extrusion type, and $(\mathbf{v}_j, h_j)$ represent the extrusion direction and length, respectively. 

The plane \( \pi_j \) is determined by the key cross-section slice associated with the input sketch \( \mathcal{K}_j \) and its normal vector is used to define the direction of the extrusion $\mathbf{v}_j$. The extrusion type \( t_j \) is assigned based on the nesting hierarchy of the loop \( \mathbf{L}_j \) within its corresponding cross-section slice \( \mathcal{S}_i \). The outermost loops are labeled as \textit{new} extrusions, and the label alternates with each subsequent level of nesting: a loop contained within a \textit{new} extrusion loop is labeled as \textit{cut} extrusion, a loop within a \textit{cut} extrusion loop is labeled as \textit{new} extrusion, and this alternating pattern continues with increasing depth. For loops labeled as \textit{new}, the corresponding length $h_{j}$ is determined through an extrusion optimization process. Loops labeled as \textit{cut} are interpreted as infinite cuts.

\vspace{-0.1cm}
 \paragraph{Extrusion Length Optimization.} To optimize over the extrusion length $h_{j}$, we define a set of \textit{extrusion vectors} over the sketch \( \mathcal{K}_{j} \) by sampling a set of $n_r$ 3D anchor points \( \{ \mathbf{r}_{k} \}_{k=1}^{n_r} \in \mathbb{R}^3 \) along the loop boundary. Each anchor point is associated with a learnable scalar extrusion length \( h_{j} \in \mathbb{R} \), shared across the anchor points of the same loop. Each anchor point $\mathbf{r}_{k}$ is mapped to an extrusion vector $\rho_k \in \mathbb{R}^3$ along the direction of the extrusion $\mathbf{v}_j$ with a length $h_{j}$ using the following function,

 \begin{equation}
    \mathbf{F} : \mathbb{R}^3 \times  \mathbb{R}^3 \rightarrow \mathbb{R}^3,\quad \mathbf{F}(\mathbf{r}_k,  h_{j}\mathbf{v}_j) :=  \mathbf{r}_k + h_{j} \mathbf{v}_j =\rho_k \ .
\end{equation}
 
 This results in a set of extrusion vectors  \( \{ \rho_{k} \}_{k=1}^{n_r} \), each of them has its corresponding anchor point $\mathbf{r}_k$ as starting point and all of them sharing the same length and direction $h_{j}$ and $\mathbf{v}_j$.  
Next, a set of $n_M$ points \( \mathcal{Q} = \{ \mathbf{q}_l \}_{l=1}^{n_M} \in \mathbb{R}^3 \) are sampled on the input mesh $\mathbf{M}$ and the set of extrusion vectors $\{\rho_u\}_{u=1}^{n_r+n_L+n_{key}}$ from each loop and each key cross-section slice are considered. The \textit{point-to-vector} distance \( d(\mathbf{q}_l, \rho_u) \) is then computed between each point $\mathbf{q}_l$ and extrusion vector $\rho_u$ to identify the closest extrusion vector    $ \rho_{\text{min}} := \arg\min_{u} \ d(\mathbf{q}_l, \rho_u)$.

The total loss is defined as the mean squared distance over all sampled points, regularized by the squared sum of extrusion lengths to avoid trivial solution. It is given by, 
\begin{equation}
    \mathcal{L}_{\text{extr}} = \frac{1}{{n_M}} \sum_{l=1}^{n_M} d(\mathbf{q}_l, \rho_{\text{min}})^2 + \lambda \sum_{i,j} h_{j}^2 \ ,
\end{equation}

\noindent where $\lambda$ denotes a scaling factor for the regularization term, $i$ refers to the detected key slice plane and $j$ refers to the individual sketch loop. The extrusion lengths \( h_{j} \) are optimized jointly for all the sketch loops over the detected key slice planes via gradient descent to minimize \( \mathcal{L}_{\text{extr}} \), thereby adjusting the extrusion vectors $\rho_u$ to best fit to the input mesh geometry.

\section{Experiments}
\label{sec:experiments}

\begin{figure}[t]
    \centering
    \includegraphics[width=\linewidth]{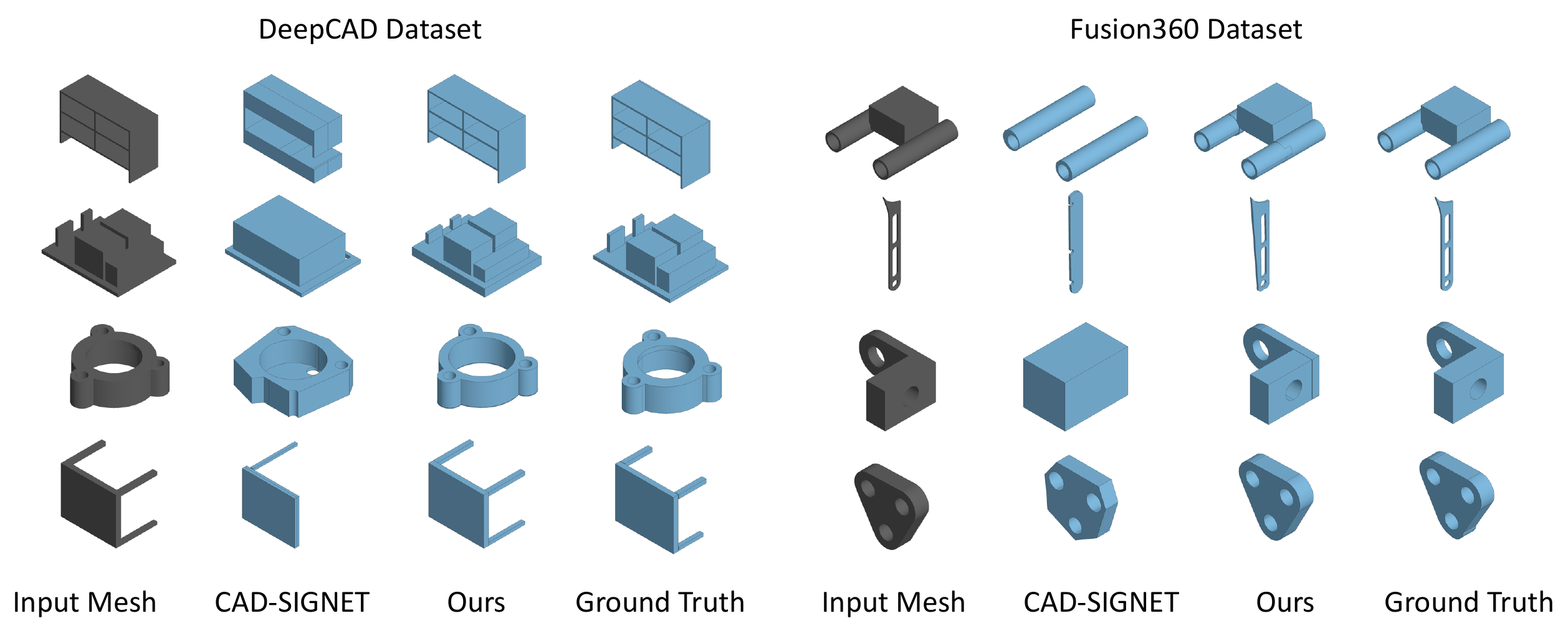}
    \caption{Qualitative comparison of of our method and that of~\cite{khan2024cad} on DeepCAD and Fusion360.}
    \label{fig:comparison}
\end{figure}

\begin{figure}[t]
    \centering
    \includegraphics[width=.9\linewidth]{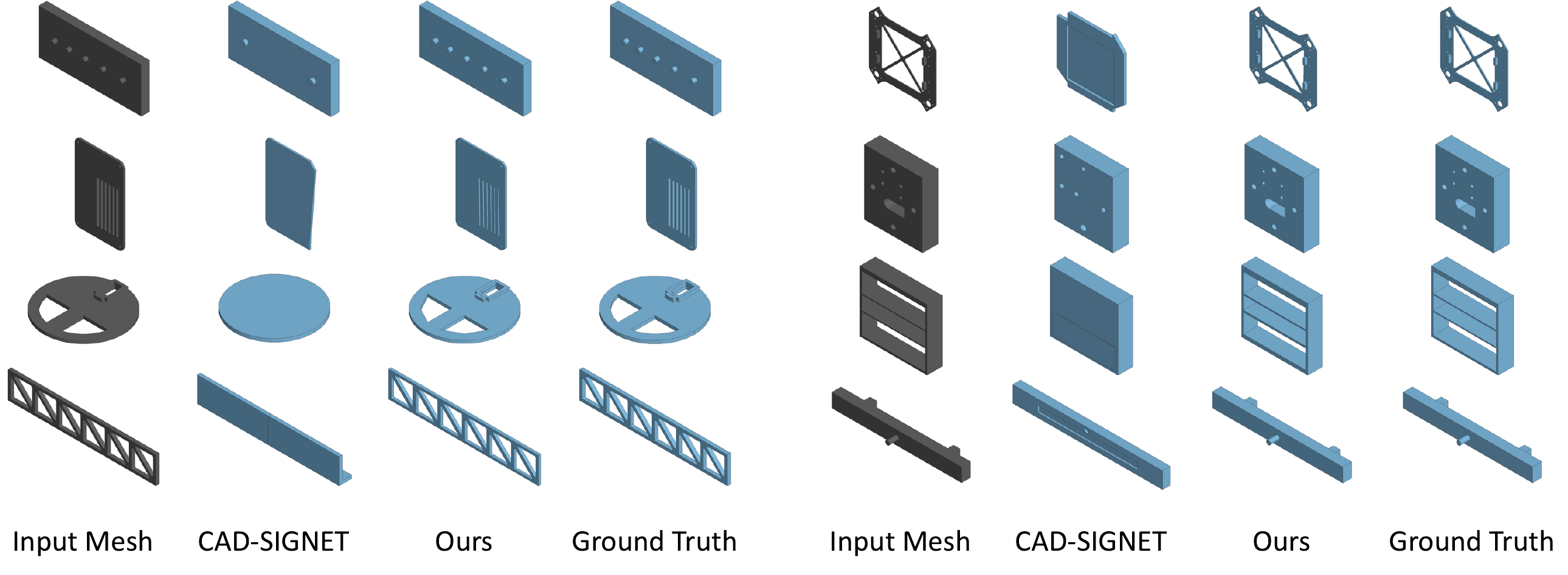}
    \caption{Qualitative comparison between our method and that of~\cite{khan2024cad} on complex models containing fine-grained geometric details.}
    \label{fig:finegrained}
\end{figure}

In this section, we detail the experimental setup and report results to evaluate the effectiveness of the proposed \modelname.

\vspace{-0.3cm}

\paragraph{Datasets.} We evaluate our method on the test sets of  DeepCAD~\cite{wu2021deepcad} and Fusion360~\cite{willis2020fusion} datasets. The sketch plane detection is trained on the train set of DeepCAD~\cite{wu2021deepcad}, while the constrained sketch parameterization network is trained on the train set of the SketchGraphs~\cite{para2021sketchgen} dataset and finetuned on an augmented version of the dataset. Details on the processing of the datasets are in the supplementary.

\vspace{-0.3cm}

\paragraph{Metrics.} For 3D CAD reconstruction, median Chamfer Distance, Intersection over Union (IoU), median Edge Chamfer Distance (ECD) and Invalidity Ratio (IR) are used. For 2D sketches, image-level sketch chamfer distance (SCD) is used. Additional metric details are provided in the supplementary.

\vspace{-0.3cm}

\paragraph{Implementation Details.} Planes are preprocessed to ensure consistent normal directions along the positive axes. The plane detection model is trained for 20 epochs on DeepCAD~\cite{wu2021deepcad} with $l_r=1 \times 10^{-4}$. The constrained sketch parameterization model is trained on SketchGraphs~\cite{para2021sketchgen} as in~\cite{karadeniz2024davinci} and finetuned for 50 epochs on synthetically generated, noise-augmented loops. We use 40 cross-sections per axis, each normalized to an image of size $128 \times 128$. Normalization is performed using 2D offsets and a scale factor corresponding to a unit bounding box. The transformer encoder of the sketch plane detection comprises 4 layers and 4 attention heads, with an embedding dimension of 256. The sketch parameterization network architecture is similar to~\cite{karadeniz2024davinci}. For the extrusion optimization, we run 200 iterations with a learning rate of $2 \times 10^{-4}$.  AdamW is used as an optimizer for all the experiments. More details are in the supplementary.

\subsection{Comparison with state-of-the-art}

\begin{table*}[!t]
\setlength{\belowcaptionskip}{-0.9cm}
\caption{Quantitative Results on DeepCAD and Fusion360 Datasets.}
\vspace{-0.2cm}
\centering
\begin{tabular}{lcccccccccc}
\toprule
\multirow{2}{*}{Method} & \multicolumn{4}{c}{DeepCAD Test Set} & \multicolumn{4}{c}{Fusion360 Test Set} \\ & Med. CD\textdownarrow & IoU\textuparrow & IR\textdownarrow & ECD\textdownarrow & Med. CD\textdownarrow & IoU\textuparrow & IR\textdownarrow & ECD\textdownarrow  \\
\midrule
DeepCAD~\cite{wu2021deepcad} & $9.64$ & $46.7$ & $7.1$ & -- & $89.2$ & $39.9$ & $25.2$ & -- \\
Point2Cyl~\cite{uy2022point2cyl} & $4.27$ & $73.8$ & $3.9$ & -- & $4.18$ & $67.5$ & $3.2$ & -- &\\
CAD-Diffuser~\cite{ma2024draw} & $3.02$ & $74.3$ & $1.5$ & -- & $3.85$ & $63.2$ & $1.7$ & -- \\
CAD-SIGNet~\cite{khan2024cad} & $0.28$
& $77.6$ & \textbf{0.9} & 0.74 & 0.56 & $65.6$ & \textbf{1.6} & 4.14 \\
Ours  & \textbf{0.20} & \textbf{80.6} & 2.6  & \textbf{0.46} & \textbf{0.48} & \textbf{68.7} & 3.2 & \textbf{2.66}\\
\bottomrule
\end{tabular}
\label{tab:comparison}
\end{table*}

The proposed method is compared with DeepCAD~\cite{wu2021deepcad}, CAD-Diffuser~\cite{ma2024draw}, Point2Cyl~\cite{uy2022point2cyl}, and CAD-SIGNet~\cite{khan2024cad} on the DeepCAD and Fusion360 test sets. A quantitative comparison with these methods is provided in Table~\ref{tab:comparison}. Across datasets, \texttt{MiCADAngelo} achieves superior reconstruction performance. Note that low IR reported by the best performing~\cite{khan2024cad} are enabled by test-time sampling, where multiple CAD reconstruction candidates are generated and the best is selected. Without test-time sampling, \cite{khan2024cad} yields an IR of $4.4$ and $9.3$ for DeepCAD and Fusion360, respectively. Figure~\ref{fig:comparison} provides a qualitative comparison with \cite{khan2024cad}.

\begin{table}[h]
\centering
\setlength{\tabcolsep}{5pt}
\caption{Quantitative results on complex models and on models with more than two extrusions.}
\label{tab:comparison_finegrained}
\begin{tabular}{lccccccccc}
\toprule
\multirow{2}{*}{Method} 
& \multicolumn{4}{c}{Models with $\geq$ 4 Loops} 
& & \multicolumn{4}{c}{Models with $>2$ Extrusions} \\
\cmidrule(lr){2-5} \cmidrule(lr){7-10}
& Med. CD$\downarrow$ & IoU$\uparrow$ & IR$\downarrow$ & ECD$\downarrow$
& & Med. CD$\downarrow$  & IoU$\uparrow$ & IR$\downarrow$ & ECD$\downarrow$ \\
\midrule
CAD-SIGNet~\cite{khan2024cad} 
& 1.34 & 49.2 & \textbf{3.2} & 4.75 
& & 3.95 & 40.6  & 5.4 & 9.81 \\
Ours 
& \textbf{0.37} & \textbf{68.3} & 4.1 & \textbf{2.04}  & & \textbf{0.46} & \textbf{64.8} & \textbf{3.0} & \textbf{2.27} \\
\bottomrule
\end{tabular}
\end{table}

Our method consistently generates CAD models that closely resemble the ground-truth geometry across a diverse range of samples. We also evaluate the performance of our method against~\cite{khan2024cad} on a subset of complex models with fine-grained details~(containing 4 or more loops) and models with more than 2 extrusions from DeepCAD. Results in Table~\ref{tab:comparison_finegrained} demonstrate that our methods significantly outperforms~\cite{khan2024cad} in preserving
fine-grained geometric details and handling complex geometries. More qualitative results on complex models are provided in Figure~\ref{fig:finegrained}.

\subsection{Impact of Constraints in 3D Reverse Engineering}
\vspace{-0.2cm}

\begin{wraptable}{r}{0.5\textwidth}
\centering
\caption{Quantitative results of deformation robustness under sketch modifications}
\setlength{\tabcolsep}{2pt}
\begin{tabular}{lccccc}
\toprule
Method & Med. CD\textdownarrow & IoU\textuparrow & IR\textdownarrow & ECD\textdownarrow \\
\midrule
    CAD-SIGNet~\cite{khan2024cad} & 2.89 & 57.4 & \textbf{3.5} & 20.43 \\
    Ours    & \textbf{0.38} & \textbf{81.1} & 4.3 & \textbf{1.29}\\
\bottomrule
\end{tabular}
    \label{tab:comparison_cpt}
\end{wraptable}

\texttt{MiCADAngelo} is the first CAD reverse engineering method to generate parametric constraints. The proposed explicit modeling of the reverse engineering pipeline enables sketch constraint inference directly from rasterized sketch images, significantly reducing the solution space compared to predicting constraints from point clouds. To demonstrate the importance of constraint modeling in 3D CAD reconstruction, we conduct an experiment analyzing how CAD reconstructions produced by \texttt{MiCADAngelo} and~\cite{khan2024cad} respond to sketch modifications similar to those typically performed by CAD designers. Specifically, we introduce a small random displacement to a point in the recovered sketch geometry and evaluate how this change affects the resulting solid geometry, comparing it to a constrained ground-truth model modified with the same displacement. Due to unavailability of CAD constraints in DeepCAD and Fusion360 datasets, we form data for this experiment by extruding $1k$ sketches with closed loops from the SketchGraphs dataset~\cite{seff2020sketchgraphs}. The impact of the displacement on the solid geometry is computed by integration with the FreeCAD API~\cite{FreeCAD}. 

\begin{figure}[t]
    \centering
    \includegraphics[width=.88\linewidth]{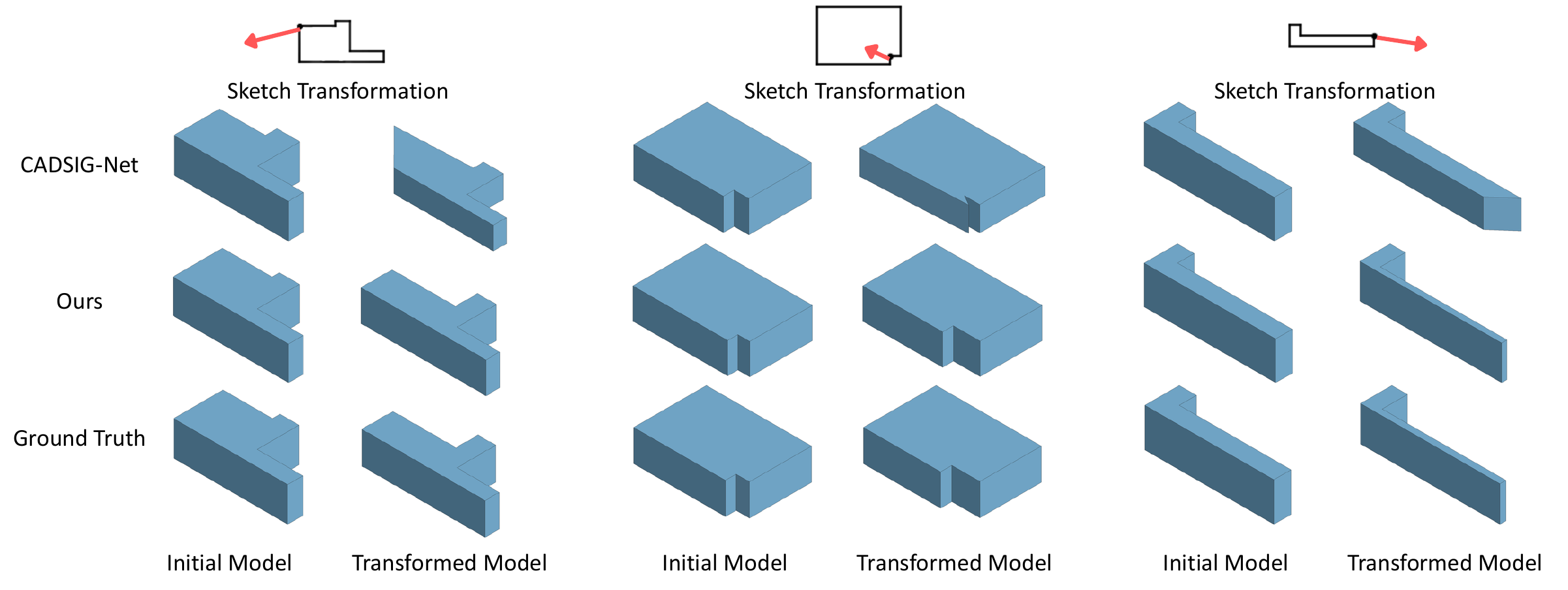}
    \caption{Qualitative comparison of the impact of the constraints by transformations applied on sketch points.}
    \label{fig:cpt_qualitative}
\end{figure}

We compare our method to that of ~\cite{khan2024cad}. Note that the output representation of~\cite{khan2024cad} generates closed sketch loops by construction, which implicitly enforces coincident constraints between sketch vertices. Results are reported in Table~\ref{tab:comparison_cpt}. \texttt{MiCADAngelo} reverse-engineered models closely match the ground truth, even after random sketch modifications. As illustrated in Figure~\ref{fig:cpt_qualitative}, due to effective application of CAD sketch constraints, edits in our method propagate correctly through sketch primitives, resulting in modified CAD models that retain structural consistency \textit{w.r.t} the ground-truth. In contrast, coincident constraints alone are insufficient to preserve overall geometry for~\cite{khan2024cad} with displacements leading to geometric distortions. Note that learning sketch constraints might also impact sketch parameterization and the overall CAD reconstruction (e.g., inferring orthogonality constraint can help in enforcing the parameters of the predicted lines to be orthogonal). While constraints are not strictly necessary for sketch primitive parameterization, prior work~[35] has shown that jointly learning primitives and constraints benefits both tasks through shared structural information. We leave further exploration of improved geometry with constraints to future work.

\vspace{-0.2cm}
\subsection{Additional Experiments}
\vspace{-0.2cm}
We include additional experiments conducted to further evaluate \modelname.

\begin{figure}[t]
\centering
\resizebox{0.45\linewidth}{!}{
\begin{minipage}{0.49\linewidth}
    \captionof{table}{Effect of contextual embeddings on plane detection performance.}
    \label{tab:plane_detection}
    \centering
    \setlength{\tabcolsep}{5pt}
    \begin{tabular}{cccc}
    \toprule
     Contextual Emb. & Precision & Recall & F1 \\
    \midrule
      \ding{55}  & 0.317 & 0.292  & 0.296 \\
      \ding{51}  & \textbf{0.894} & \textbf{0.864} & \textbf{0.870} \\
    \bottomrule
    \label{tab:embeddings}
    \end{tabular}

    \vspace{1em}

    \captionof{table}{Plane detection performance across datasets.}
    \centering
    \setlength{\tabcolsep}{5pt}
    \begin{tabular}{lccc}
    \toprule
    Dataset & Precision & Recall & F1 \\
    \midrule
    DeepCAD   & 0.894 & 0.864 & 0.870 \\
    Fusion360 & 0.860 & 0.812 & 0.820 \\
    CC3D      & 0.803 & 0.790 & 0.777 \\
    \bottomrule
    \label{tab:plane_generalization}
    \end{tabular}
\end{minipage}
}
\hfill
\begin{minipage}{0.54\linewidth}
\centering
\includegraphics[width=0.85\linewidth]{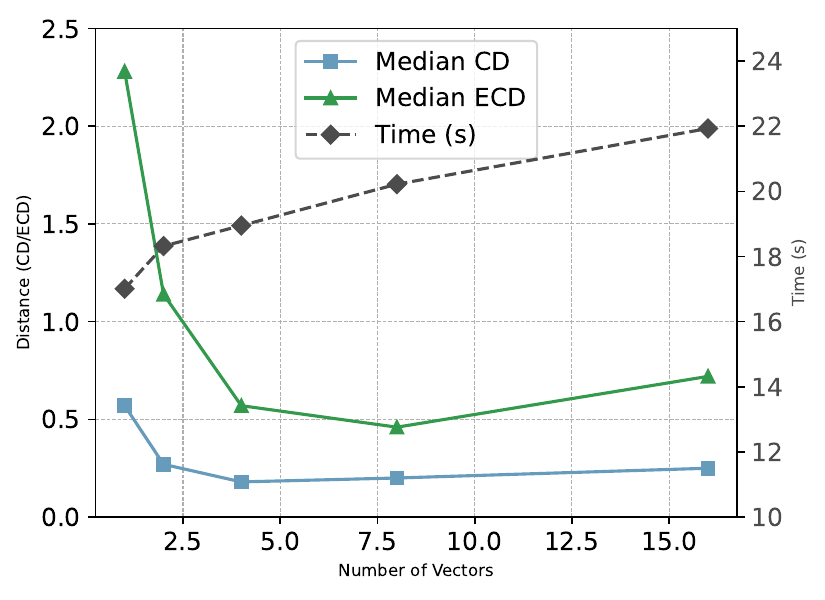}
\vspace{-0.3cm}
\caption{CD and ECD performance and inference times as a function of the number of extrusion vectors.}
\label{fig:num_vectors}
\end{minipage}
\end{figure}

\textbf{Plane Detection.} Table~\ref{tab:embeddings} ablates the effect of contextual embeddings in slice encoding. Incorporating positional information to the geometric features of the cross-section leads to improved performance. Table~\ref{tab:plane_generalization} shows robust plane detection performance with minimal drop in Fusion360 and CC3D datasets.

\begin{wraptable}{r}{0.48\textwidth}
\centering
\caption{Evaluation of sketch parameterization on single-extrusion CAD models.}
\label{tab:sketch_param}
\setlength{\tabcolsep}{20pt}
\begin{tabular}{lc}
\toprule
Method & \textit{Avg. SCD} \\
\midrule
Davinci~\cite{karadeniz2024davinci} & 0.827 \\
Ours & \textbf{0.283} \\
\bottomrule
\end{tabular}
\end{wraptable}

\textbf{Constrained Sketch Parameterization.} We evaluate our constrained sketch parameterization network against the recent proposed~\cite{karadeniz2024davinci}. To ensure that ground-truth sketches correspond to valid cross-sections, this experiment is conducted on single-extrusion models from the DeepCAD test set. Due to fine-tuning on synthetically generated noisy closed loops, \modelname~outperforms~\cite{karadeniz2024davinci}.

\textbf{Extrusion Optimization.} Figure~\ref{fig:num_vectors} reports differentiable extrusion performance for a varying number of vectors. Performance improves steadily up to 8 vectors, with minimal impact on inference time.

\begin{figure}[t]
\vspace{-0.2cm}
    \centering
    \includegraphics[width=.85\linewidth]{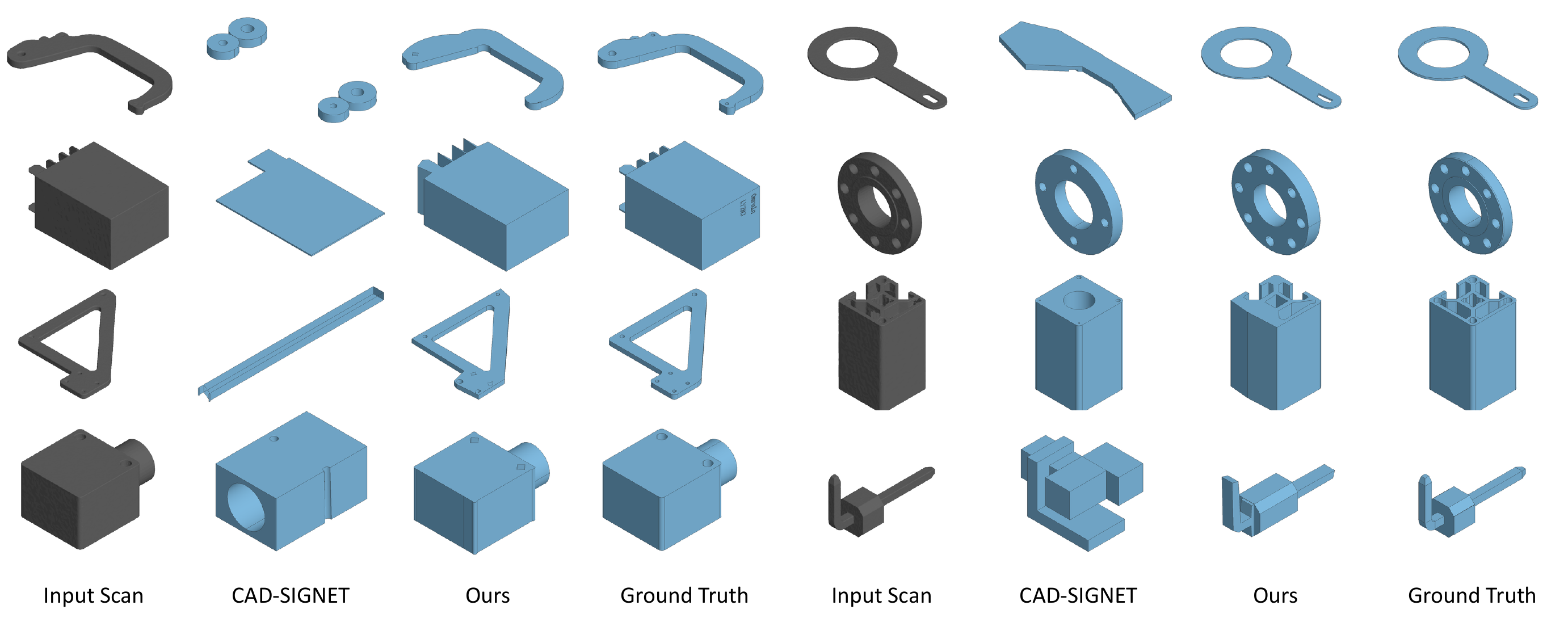}
    \caption{Qualitative comparison of the performance on the real-world scans from CC3D.}
    \label{fig:cc3d_qualitative}
\end{figure}

\textbf{Robustness to Real-world Scans.} To evaluate performance on real imperfect scans, we include a quantitative comparison on the challenging CC3D~\cite{9191095,mallis2023sharp} dataset. This cross-dataset evaluation compares our method to that of \cite{khan2024cad}, both trained on DeepCAD and evaluated on CC3D scans that include realistic scanning artifacts, such as holes and misoriented normals. Results are shown in the Table~\ref{tab:cc3d}. The proposed MiCADangelo outperforms \cite{khan2024cad} in this in-the-wild setting, demonstrating greater robustness to real-world noise. Figure~\ref{fig:cc3d_qualitative} presents a qualitative evaluation on real-world 3D scans from the CC3D~\cite{9191095,mallis2023sharp} dataset. The results demonstrate the robustness of our approach in handling real-world artifacts and producing cleaner, more accurate CAD models compared to~\cite{khan2024cad}. 

\begin{table}[h]
\centering
\caption{Comparison between CADSIG-Net and our method on CC3D dataset.}
\begin{tabular}{lccccc}
\toprule
Method & Med. CD\textdownarrow & IoU\textuparrow & IR\textdownarrow & ECD\textdownarrow\\
\midrule
CADSIG-Net~\cite{khan2024cad} & 2.90 & 42.6  & 4.4 & 8.68 \\
Ours       & \textbf{1.69} & \textbf{ 50.8} &  \textbf{2.2} &  \textbf{5.93} \\
\bottomrule
\end{tabular}
\label{tab:cc3d}
\end{table}

\vspace{-0.3cm}
\section{Conclusion}
\label{sec:conclusion}
\vspace{-0.3cm}

In this work we proposed \texttt{MiCADAngelo}, a novel CAD reverse engineering approach that transforms 3D scans into fully parametric CAD models. Our method is inspired by real-world CAD practices and uniquely incorporates cross-sections and sketch primitives with constraints, enabling the preservation of both high-level parametric structure and fine-grained geometric detail. Evaluation is performed on standard benchmarks were our approach outperforms existing methods establishing a new state-of-the-art in CAD model reconstruction from 3D scans.

\textbf{Limitations}. A detailed discussion on failure cases and limitations is provided in supplementary. Among CAD operations, the proposed method only supports extrusion as in previous works~\cite{khan2024cad,li2023secad,uy2022point2cyl}. Extrusions are currently defined based on sketch plane normals, which can be suboptimal for models with non-axis-aligned extrusions. Note that such complex models are also challenging for~\cite{wu2021deepcad, ma2024draw, khan2024cad}. The current method does not yet support complex sketch primitives such as B-splines. Future works will address these limitations.

\section{Acknowledgements}
This work is supported by the National Research Fund (FNR), Luxembourg, under the BRIDGES2021/IS/16849599/FREE-3D project and by Artec3D. 

{
    \small
    \bibliographystyle{unsrtnat}
    \bibliography{main}
}

\title{\texttt{MiCADangelo}: Fine-Grained Reconstruction of Constrained CAD Models from 3D Scans}

\maketitlesupplementary

\subsection*{Overview}

This supplementary document provides additional technical details complementing the main paper, along with an extended set of qualitative evaluations of the proposed \texttt{MiCADangelo}. Section~\ref{sec:data_preprocess} outlines the data preprocessing procedures, Section~\ref{sec:metrics} describes the evaluation metrics, Section~\ref{sec:additional_exp} provides additional experimental analysis, Section~\ref{sec:limitations} discusses current limitations and failure cases, and Section~\ref{sec:more_qualitative} presents additional qualitative results.

\subsection{Data Preprocessing}
\label{sec:data_preprocess}

This section describes the data preparation steps used for training and evaluation of the proposed method.

\subsubsection{Data Preparation for the Training of Plane Detection Network}

\textbf{Preprocessing of extrusion planes.} To train the proposed plane detection network, extrusion planes are extracted from the DeepCAD~\cite{wu2021deepcad} dataset using the original train and test splits.
For every extrusion in a CAD sequence, we start from its sketch plane
\((\mathbf{o},\mathbf{n})\) with origin $\mathbf{o} \in \mathbb{R}^3$ and normal $\mathbf{n} \in \mathbb{R}^3$, forward extent \(e_1 \in \mathbb{R}\) and backward extent \(e_2 \in \mathbb{R}\).
We compute a single, canonically oriented plane
\((\mathbf{o}^{\star},\mathbf{n}^{\star})\). First, the origin $\mathbf{o}$ is moved depending on the
      \texttt{extent\_type} flag provided by the CAD sequence:
      \[
      \mathbf{o}^{\star}=
      \begin{cases}
      \mathbf{o}-e_{1}\mathbf{n}, & \text{if } \texttt{extent\_type} = \text{symmetric extrusion },\\[2pt]
      \mathbf{o}-e_{2}\mathbf{n}, & \text{if } \texttt{extent\_type} = \text{two-sided extrusion },\\[2pt]
      \mathbf{o}, & \text{if } \texttt{extent\_type} = \text{one-sided extrusion }.
      \end{cases}
      \]
The two limiting points are
      \(\mathbf{o}_{\mathrm{fwd}}=\mathbf{o}+e_{1}\mathbf{n}\) and
      \(\mathbf{o}_{\mathrm{bwd}}=\mathbf{o}+e_{2}\mathbf{n}\).
      The normalized connecting vector \(\mathbf{n}' = (\mathbf{o}_{\mathrm{fwd}} - \mathbf{o}_{\mathrm{bwd}})/\|\mathbf{o}_{\mathrm{fwd}} - \mathbf{o}_{\mathrm{bwd}}\|\) is used as the normal direction.
 To make the direction unambiguous, we flip it towards the positive axis $
      \mathbf{n}^{\star}=|\mathbf{n}'|
      $. If the direction is flipped, $\mathbf{o}^{\star}$ is moved to the point $\mathbf{o}_{\mathrm{bwd}}$.
Finally, the set \(\{(\mathbf{o}^{\star}_{j},\mathbf{n}^{\star}_{j})\}_{j=1}^{n_s}\) where $n_s$ is the number of extrusions,
constitutes the ground-truth extrusion planes.

\textbf{Preparation of Slicing Plane Labels.}  
For each CAD model in the DeepCAD dataset, we sample equally spaced \(40\) slicing planes along each of the \(x\)-, \(y\)-, and \(z\)-axes, producing a total of \(N = 120\) slicing planes:
\[
\Pi = \bigl\{ \pi_{a,\sigma} \;\big|\; a \in \{x, y, z\},\; \sigma \in \{0, \ldots, 39\} \bigr\}.
\]
We define a binary label set \(\mathbf{y} = \{y_i\}_{\pi_i \in \Pi}\), where \(y_i \in \{0, 1\}\)\ for each slicing plane \(\pi_i\), and initialize all labels to \(0\).  
For every ground-truth plane \(\pi_k\), we identify the nearest slicing plane:
\[
\pi_k^{\star} = \arg\min_{\pi_i \in \Pi} d(\pi_i, \pi_k),
\]
and update the corresponding label \(y_k\) to \(1\), where \(d(\pi_i, \pi_k)\) denotes the distance between planes \(\pi_i\) and \(\pi_k\).

\subsubsection{Data Preparation for the Training of Sketch Parameterization Network}

\begin{figure}[h]
    \centering
        \setlength{\belowcaptionskip}{-0.1cm}
    \includegraphics[width=\linewidth]{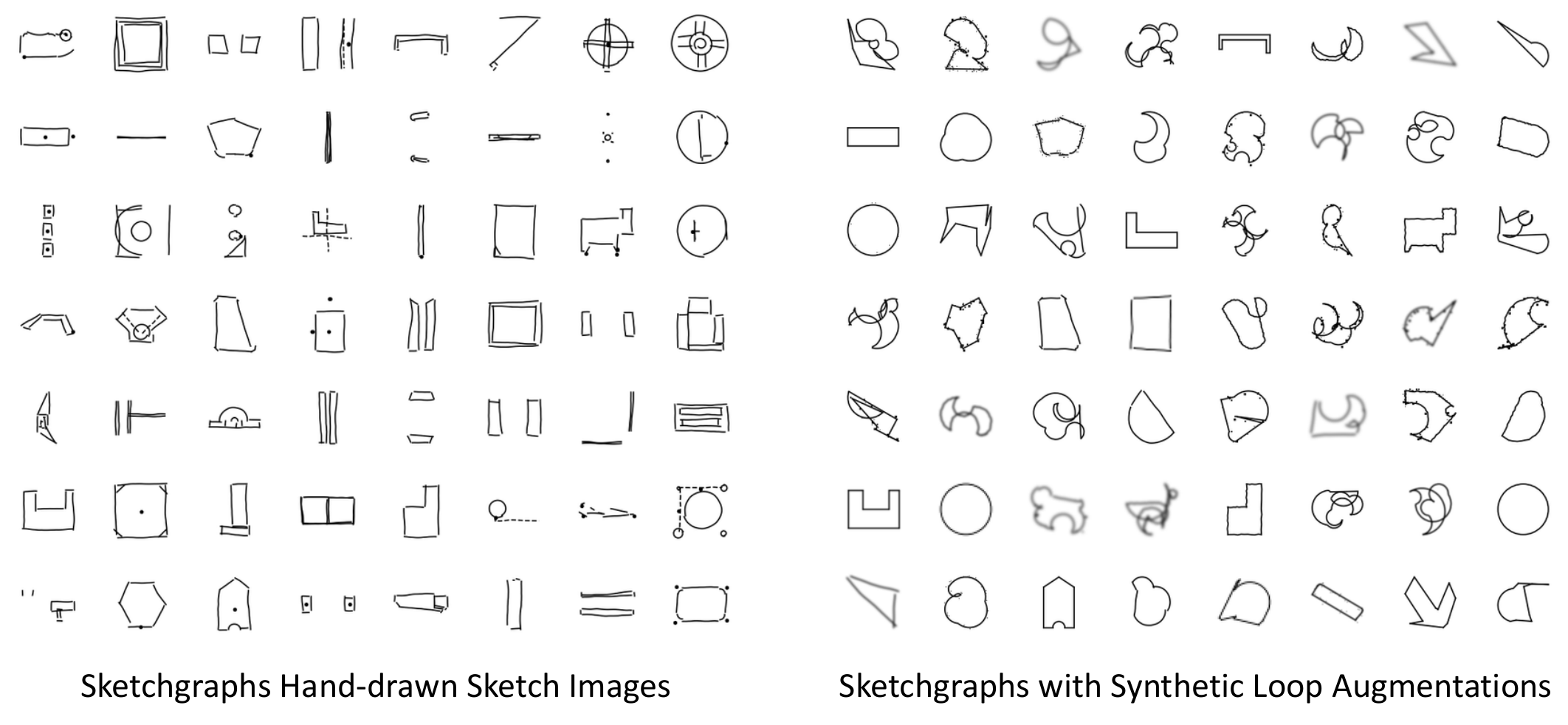}
        \vspace{-0.5cm}
    \caption{Examples of original (left) and augmented SketchGraphs sketches (right), illustrating the effect of synthetic loop generation and noisy rendering.}
    \label{fig:sketchgraphs_aug}
\end{figure}

The constrained sketch parameterization network is initially trained on the SketchGraphs dataset~\cite{seff2020sketchgraphs}. It is then fine-tuned on an augmented version of SketchGraphs that includes added noise and synthetically generated closed-loop images. This augmentation helps the model better adapt to the characteristics of cross-sectional slice data. During training, a synthetic sketch is used if a uniformly sampled random number exceeds 0.5; otherwise, the original SketchGraphs sketch is retained. A random rotation between $[0, 2\pi]$ is also applied to the sketch with random probability $0.2$. The procedures for generating random sketches and noise augmentation are detailed in Algorithm~\ref{alg:gen_random_sketch} and Algorithm~\ref{alg:add_noise}, respectively. Figure~\ref{fig:sketchgraphs_aug} shows example sketches from the original SketchGraphs dataset and its augmented counterpart, highlighting the added variability introduced by synthetic loops and image-space noise.

\begin{algorithm}
\caption{GenerateRandomLoopSketch}
\begin{algorithmic}[1]
\Require $\text{maxPrimitives}$, $\text{arcWeight}$
\Ensure $\mathcal{K} = (\mathcal{P}, \mathcal{C})$ \Comment{Sketch primitives $\mathcal{P}$ and constraints $\mathcal{C}$}
\State $\mathcal{P} \gets \emptyset$, $\mathcal{C} \gets \emptyset$
\State $n_p \gets \text{random\_int}(3, \text{maxPrimitives})$
\State $P \gets \text{RandomPolygon}(n_p)$\footnotemark  \Comment{Random polygon with at least 3 vertices}

\For{$i = 0$ to $n_p - 1$}
    \State $(\mathbf{x}_s, \mathbf{x}_e) \gets$ edge $i$ of $P$
    \If{$\text{rand()} > \text{arcWeight}$}
        \State $\mathbf{p}_i \gets \text{Line}(\mathbf{x}_s, \mathbf{x}_e)$ \Comment{Create line primitive}
    \Else
\State $\mathbf{x}_m \gets \frac{1}{2}(\mathbf{x}_s + \mathbf{x}_e)$ \Comment{Midpoint of the start and end points}
        \State $\mathbf{v} \gets \mathbf{x}_e - \mathbf{x}_s$ %
        \State $\delta \gets \text{rand\_uniform}(0.05, 0.15)$
        \State $\mathbf{x}_a \gets \mathbf{x}_m + \delta \cdot \mathbf{v}_\perp$ \Comment{Add offset to midpoint}
        \If{$\text{rand()} > 0.8$}
            \State swap $\mathbf{x}_s \leftrightarrow \mathbf{x}_e$ \Comment{Randomly reverse arc direction}
        \EndIf
        \State $\mathbf{p}_i \gets \text{Arc}(\mathbf{x}_s, \mathbf{x}_a, \mathbf{x}_e)$ \Comment{Create arc primitive}
    \EndIf

    \State $\mathcal{P} \gets \mathcal{P} \cup \{\mathbf{p}_i\}$ \Comment{Add primitive to sketch}

    \If{$i > 0$}
        \State $\mathbf{c}_{i} \gets (\mathbf{p}_{i-1},\; \mathbf{p}_i,\; \text{coincident})$ \Comment{Add coincident constraint}
        \State $\mathcal{C} \gets \mathcal{C} \cup \{\mathbf{c}_i\}$
    \EndIf
\EndFor

\State $\mathbf{c}_{n_p} \gets (\mathbf{p}_{n_p-1},\; \mathbf{p}_0,\; \text{coincident})$ \Comment{Close loop with final constraint}
\State $\mathcal{C} \gets \mathcal{C} \cup \{\mathbf{c}_{n_p}\}$

\State \Return $\mathcal{K} = (\mathcal{P}, \mathcal{C})$
\end{algorithmic}
\label{alg:gen_random_sketch}
\end{algorithm}
\footnotetext{Implemented using \href{https://github.com/bast/polygenerator}{polygenerator} and \href{https://github.com/shapely/shapely}{shapely} libraries.}

\begin{algorithm}
\caption{RenderWithNoiseAugmentation}
\begin{algorithmic}[1]
\Require Sketch $\mathcal{K}$, \texttt{is\_hand\_drawn}
\Ensure Sketch image $\mathbf{X}$
\State $s \gets 128$
\If{$\text{rand()} < 0.2$}
    \State $s \gets \text{rand\_choice}(\{64,\; 128,\; 256\})$ \Comment{Random rescaling size}
\EndIf
\State $\mathbf{X} \gets \text{RenderSketch}(\mathcal{K},\; \texttt{is\_hand\_drawn},\; s)$ \Comment{Render sketch at resolution $s$}
\State $\mathbf{X} \gets \text{Resize}(\mathbf{X},\; 128 \times 128)$ \Comment{Normalize size}
\If{$\text{rand()} < 0.2$}
    \State $\mathbf{X} \gets \text{AddNoiseNearForeground}(\mathbf{X})$ \Comment{Add local noise}
\EndIf
\If{$\text{rand()} < 0.2$}
    \State $k \gets \text{rand\_choice}(\{3,\; 5\})$
    \State $d \gets 2k + 1$
    \State $\mathbf{X} \gets \text{GaussianBlur}(\mathbf{X},\; \text{kernel\_size} = d,\; \sigma = 0)$ \Comment{Apply blur with random kernel size}
\EndIf
\State \Return $\mathbf{X}$
\end{algorithmic}
\label{alg:add_noise}
\end{algorithm}

\vspace{1em}

\begin{algorithm}
\caption{AddNoiseNearForeground}
\begin{algorithmic}[1]
\Require Image $\mathbf{X}$ of size $H \times W$, max offset $d$
\Ensure Modified image $\mathbf{X}$
\State $B \gets \{(x, y)\;|\; \mathbf{X}[y, x] = 0\}$ \Comment{Foreground (black) pixel coordinates}
\If{$|B| = 0$}
    \State \Return $\mathbf{X}$
\EndIf
\State $N \gets \text{randint}(0,\; 100)$ \Comment{Number of noise points}
\For{$i = 1$ to $N$}
    \State $(x, y) \sim \text{Uniform}(B)$ \Comment{Sample uniformly from foreground}
    \State $\delta_x, \delta_y \gets \text{randint}(-d,\; d)$
    \State $x' \gets \text{clip}(x + \delta_x,\; 0,\; W-1)$
    \State $y' \gets \text{clip}(y + \delta_y,\; 0,\; H-1)$ \Comment{Clip to bounds}
    \If{$\text{rand()} < 0.5$}
        \State $\mathbf{X}[y', x'] \gets 0$ \Comment{Add synthetic black pixel}
    \Else
        \State $\mathbf{X}[y', x'] \gets \text{randint}(0,\; 20)$ \Comment{Add light gray speckle}
    \EndIf
\EndFor
\State \Return $\mathbf{X}$
\end{algorithmic}
\end{algorithm}

\subsubsection{Data Preparation for the Experimental Evaluation}

\textbf{DeepCAD Complex Subset.}
To assess the performance of the proposed method on more complex cases (Section 5.1 of the main paper), we selected a subset of CAD models from the DeepCAD test set based on the number of sketch loops. While a model can be complex even with a single loop, the DeepCAD dataset contains a disproportionately large number of very simple designs (e.g., cubes) that can dominate the overall evaluation. To mitigate this, we filtered for models containing more than four distinct loops. This process yielded a subset of 1,391 samples used for targeted evaluation.

\textbf{Extruded SketchGraphs Dataset.} To evaluate the impact of constraints in 3D reverse engineering (Section 5.2 of the main paper), we construct a set of solids with their corresponding constrained sketches from the Sketchgraphs~\cite{seff2020sketchgraphs} dataset. Starting from the first sketch in the test split, we iterate sequentially through the dataset and process only sketches that form closed loops.  Each closed sketch is loaded into FreeCAD~\cite{FreeCAD}, placed at the origin with its normal aligned to \(+z\), and extruded by \(0.3\) units along that axis to produce a 3D solid.  The solid is then exported in three formats: \texttt{OBJ}, \texttt{STEP}, and native \texttt{FCStd}. This process is repeated until 1,000 solids are generated.

\textbf{Constrained Sketch Transformations.}  
As discussed in Section~5.2 of the main paper, the impact of geometric constraints is evaluated by applying controlled transformations to the sketches in the \textit{Extruded SketchGraphs} dataset. For each sketch, we randomly displace a single point, triggering a constraint-driven update to the overall geometry. This results in a modified sketch and a corresponding transformed 3D shape. In total, this procedure yields 1,000 CAD models with associated transformations. To enable applying the same transformation to predicted sketches, we record the original point and its displacement vector during transformation generation. In the predicted sketch, we identify the closest point—among the start, midpoint, or end of all primitives—to the originally displaced point, and apply the same displacement vector. Examples of such transformations are shown in Figure~5 of the main paper and Figure~\ref{fig:transformations} in the supplementary.

\textbf{DeepCAD Cross-section Images.}
To evaluate the performance of the sketch parameterization network on cross-sectional data (Section 5.3 of the main paper), a total of 5,106 cross-section images are extracted from single-extrusion models in the DeepCAD test set. Each cross-section image is generated by slicing the corresponding 3D mesh along the sketch plane specified in the associated CAD sequence.

\begin{figure}[t]
    \centering
    \includegraphics[width=0.8\linewidth]{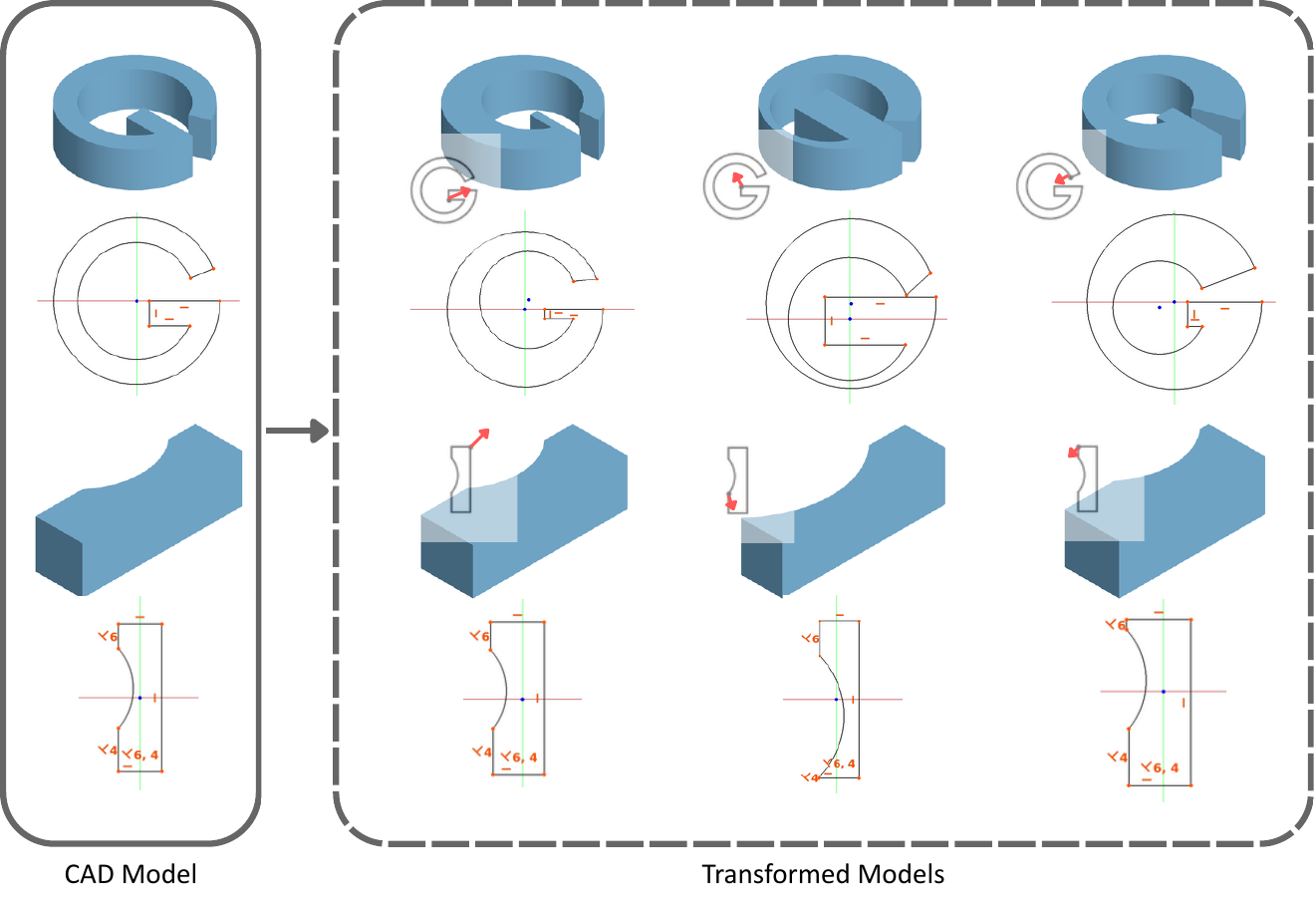}
    \vspace{-0.3cm}
    \caption{Examples of constrained sketch transformations and their effect on the resulting 3D models.}
    
    \label{fig:transformations}
\end{figure}

\subsection{Metrics}
\label{sec:metrics}

Quantitative evaluation of the proposed method was conducted using the metrics described below.

\textbf{Chamfer Distance (CD).} Let $\mathbf{M}_\text{pred}$ and $\mathbf{M}_\text{gt}$ denote the meshes obtained from the predicted and ground-truth CAD models, respectively. A fixed number of points are uniformly sampled from the surface of each mesh, yielding two point sets:
\[
\mathbb{P} = \{\mathbf{p}_i\}_{i=1}^{N}, \quad \hat{\mathbb{P}} = \{\hat{\mathbf{p}}_i\}_{i=1}^{N},
\]
where $N=8192$ and $\mathbf{p}_i, \hat{\mathbf{p}}_i \in \mathbb{R}^3$. Sampled points $\mathbb{P}$ and $\hat{\mathbb{P}}$ are normalized to unit scale. The \textit{Chamfer Distance} between the sampled point sets is then defined as:
\begin{equation}
\operatorname{CD} = \frac{1}{2N} \sum_{i=1}^{N} \min_{\hat{\mathbf{p}}_j \in \hat{\mathbb{P}}} \| \mathbf{p}_i - \hat{\mathbf{p}}_j \|_2^2 
+ \frac{1}{2N} \sum_{j=1}^{N} \min_{\mathbf{p}_i \in \mathbb{P}} \| \hat{\mathbf{p}}_j - \mathbf{p}_i \|_2^2.
\label{eq:chamfer3d}
\end{equation}
\textbf{Intersection over Union (IoU).} To evaluate the volumetric similarity between two 3D meshes, we compute the \textit{Intersection over Union (IoU)}. Let $\mathbf{M}_{\text{pred}}$ and $\mathbf{M}_{\text{gt}}$ be the predicted and ground-truth meshes obtained from corresponding CAD models, each possibly composed of multiple connected components $
\mathbf{M}_{\text{pred}} = \bigcup_{i} \mathbf{M}^{(i)}_{\text{pred}}, \quad 
\mathbf{M}_{\text{gt}} = \bigcup_{j} \mathbf{M}^{(j)}_{\text{gt}}.
$
The \textit{IoU} is defined as

\begin{equation}
\operatorname{IoU}(\mathbf{M}_{\text{pred}}, \mathbf{M}_{\text{gt}}) \;=\;
\frac{\displaystyle\sum_{i,j} \operatorname{Vol}\!\bigl(\mathbf{M}^{(i)}_{\text{pred}}\cap \mathbf{M}^{(j)}_{\text{gt}}\bigr)}
     {\displaystyle\sum_{i} \operatorname{Vol}\!\bigl(\mathbf{M}^{(i)}_{\text{pred}}\bigr)
      \;+\;\sum_{j} \operatorname{Vol}\!\bigl(\mathbf{M}^{(j)}_{\text{gt}}\bigr)
      \;-\;\sum_{i,j} \operatorname{Vol}\!\bigl(\mathbf{M}^{(i)}_{\text{pred}}\cap \mathbf{M}^{(j)}_{\text{gt}}\bigr)} \,
\label{eq:iou_compact}
\end{equation}
where $\operatorname{Vol}(\cdot)$ denotes the volume of a mesh\footnote{The volume of a mesh is computed using \href{https://github.com/mikedh/trimesh}{trimesh} library.}.

\textbf{Edge Chamfer Distance (ECD).}  
To evaluate performance on boundary curves, we compare models using
points sampled only from their edges. For each \texttt{STEP} file obtained from prediction and ground-truth, we draw
\(M=4096\) points uniformly across all edge curves, center the point cloud at the
origin, and normalize it to the unit bounding box, yielding two sets
\[
\mathbb{B} = \{\mathbf{b}_i\}_{i=1}^{M}, \qquad
\hat{\mathbb{B}} = \{\hat{\mathbf{b}}_i\}_{i=1}^{M},
\]
where $\mathbf{b}_i, \hat{\mathbf{b}}_i \in \mathbb{R}^3$. The \emph{Edge Chamfer Distance} between the predicted and ground-truth edge
sets is then defined as:
\begin{equation}
\operatorname{ECD} =
\frac{1}{2M} \sum_{i=1}^{M}
\min_{\mathbf{b}_j \in \mathbb{B}}
\|\hat{\mathbf{b}}_i - \mathbf{b}_j\|_2^2
\;+\;
\frac{1}{2M} \sum_{i=1}^{M}
\min_{\hat{\mathbf{b}}_j \in \hat{\mathbb{B}}}
\|\mathbf{b}_i - \hat{\mathbf{b}}_j\|_2^2 .
\end{equation}

\textbf{Invalidity Ratio (IR).} The \textit{Invalidity Ratio (IR)} is defined as the percentage of predicted CAD models that fail to produce valid meshes during export.

\textbf{Sketch Chamfer Distance (SCD).} 
To assess the similarity between predicted and ground-truth sketches, we compute a 2-D Chamfer Distance between the sets of foreground pixel coordinates obtained from their $h \times w$ binary rasterizations.  
Let the sets of foreground pixel coordinates be
\[
\mathbb{S} = \{\mathbf{s}_n\}_{n=1}^{N_f}, 
\qquad
\hat{\mathbb{S}} = \{\hat{\mathbf{s}}_n\}_{n=1}^{\hat{N}_f}
\]
where $\mathbf{s}_n, \hat{\mathbf{s}}_n \in \{(i,j)\mid 1 \le i \le h,\; 1 \le j \le w\}$, and $N_f$, $\hat{N}_f$ denote the number of foreground pixels in the ground-truth and predicted rasterizations, respectively.  
The bidirectional \textit{Sketch Chamfer Distance} is then defined as
\begin{equation}
\operatorname{SCD} =
\frac{1}{2\hat{N}_f}\sum_{n=1}^{\hat{N}_f}
\min_{\mathbf{s}_k \in \mathbb{S}}
\|\hat{\mathbf{s}}_n - \mathbf{s}_k\|_2^2
\;+\;
\frac{1}{2N_f}\sum_{n=1}^{N_f}
\min_{\hat{\mathbf{s}}_k \in \hat{\mathbb{S}}}
\|\mathbf{s}_n - \hat{\mathbf{s}}_k\|_2^2 .
\label{eq:scd}
\end{equation}

\textbf{Plane Detection Metrics (Precision, Recall, F1).} To evaluate the performance of the sketch plane detection network, we treat each slicing plane across all CAD models as a binary classification instance.  Let \(y_i\in\{0,1\}\) be the ground-truth label of the
\(i\)-th slicing plane (\(y_i=1\) if the plane coincides with a sketch
plane, \(0\) otherwise) and \(\hat{y}_i\in\{0,1\}\) the corresponding
network prediction. Then, \textit{Precision}, \textit{Recall} and \textit{F1} scores are computed as follows:

\[
\text{TP}= \sum_i \mathbbm{1}\!\bigl[y_i = 1 \,\wedge\, \hat{y}_i = 1\bigr],\qquad
\text{FP}= \sum_i \mathbbm{1}\!\bigl[y_i = 0 \,\wedge\, \hat{y}_i = 1\bigr],\qquad
\text{FN}= \sum_i \mathbbm{1}\!\bigl[y_i = 1 \,\wedge\, \hat{y}_i = 0\bigr].
\]

\[
\textit{Precision}
= \frac{\text{TP}}{\text{TP}+\text{FP}},\qquad
\textit{Recall}
= \frac{\text{TP}}{\text{TP}+\text{FN}},\qquad
\textit{F1}
= \frac{2\,\text{TP}}{2\,\text{TP}+\text{FP}+\text{FN}}
= \frac{2\,\textit{Precision}\times\textit{Recall}}{\textit{Precision}+\textit{Recall}}.
\]

\clearpage

\subsection{Additional Experimental Analysis}
\label{sec:additional_exp}

This section extends the main paper’s experimental analysis with further evaluations on complex geometries, plane detection, error accumulation, and cut extrusions.

\subsubsection{Complex Geometries}

We further extend the complex geometry analysis by evaluating a subset of highly complex models containing more than eight loops (271 models). These models represent parts with multiple intricate geometric details. The quantitative comparisons on these partitions are reported in Table~\ref{tab:complex_1}.

\begin{table}[h]
\centering
\setlength{\belowcaptionskip}{-0.5cm}
\begin{tabular}{lccccc}
\toprule
Method & Mean CD\textdownarrow & Median CD\textdownarrow & IR\textdownarrow & IoU\textuparrow & ECD\textdownarrow  \\
\midrule
CADSIG-Net~\cite{khan2024cad} & 6.48 & 1.64 & \textbf{5.1} & 41.07 & 4.84 \\
Ours       & \textbf{5.47} & \textbf{0.45} & 5.5 & \textbf{64.30} & \textbf{2.12} \\
\bottomrule
\end{tabular}
\caption{Performance on models containing more than 8 loops.}
\label{tab:complex_1}
\end{table}

\subsubsection{Plane Detection}

In Table 4 of the main paper, we ablate the impact of contextual embeddings on plane detection performance. We extend this analysis to evaluate their effect on the full CAD reconstruction pipeline. The results are presented in the Table~\ref{tab:contextual_full}. We observe that contextual embeddings have an effect on the overall CAD reconstruction performance on all reported metrics. Nonetheless, it is interesting to observe that, despite significantly underperforming in cross-section plane detection, our method without contextual embeddings still achieves reasonable overall CAD reconstruction performance. This can be attributed to the robustness of the subsequent stages—sketch parameterization and extrusion optimization—which are capable of recovering plausible reverse engineering design paths, even when the detected cross-section planes deviate from the ground truth.

\begin{table}[h]
\centering
\begin{tabular}{cccccc}
\toprule
Contextual Emb. & Mean CD\textdownarrow & Median CD\textdownarrow & IR\textdownarrow & IoU\textuparrow & ECD\textdownarrow  \\
\midrule
 \ding{55} & 8.80 & 0.39 & 3.5 & 69.4 & 2.25 \\
 \ding{51} & \textbf{2.27 } & \textbf{0.20} & \textbf{2.6} & \textbf{80.6} & \textbf{0.46} \\
\bottomrule
\end{tabular}
\caption{Comparison of CAD reconstruction with and without contextual embeddings in plane detection.}
\label{tab:contextual_full}
\end{table}

The value of $N$ controls the density of cross-section slice sampling, determining the granularity of the input to the reconstruction pipeline. We conduct an ablation on DeepCAD dataset with different  values in Table~\ref{tab:ablation_N}. We observe that larger  values provide the input model with a more dense input representation and slightly increase CAD reconstruction performance across metrics. In this work, we set $N$ to 40, which offers a good balance between accuracy and computational cost. A similar compromise arises in point‑cloud pipelines (as in all competitors \cite{khan2024cad,ma2024draw,uy2022point2cyl,wu2021deepcad}), where the input cloud is routinely down‑sampled to a fixed number of points.

\begin{table}[h]
\centering
\begin{tabular}{lccccc}
\toprule
$N$ & Mean CD\textdownarrow & Median CD\textdownarrow & IR\textdownarrow & IoU\textuparrow & ECD\textdownarrow  \\
\midrule
10 & 3.20 & 0.26 & 4.5 & 78.2 & 0.54 \\
20 & 3.17 & 0.25 & 4.3 & 78.4 & 0.53 \\
40 & \textbf{2.27} & \textbf{0.20} & \textbf{2.6} & \textbf{80.6} & \textbf{0.46} \\
\bottomrule
\end{tabular}
\caption{Ablation on the number of 2D cross-sectional slices $N$ used as input to the slice plane detection network.}
\label{tab:ablation_N}
\end{table}

\subsubsection{Error Accumulation Analysis}

Error accumulation is an inherent challenge in all recent CAD reverse engineering approaches including \modelname. This is due to the fact that a CAD model is typically represented as a sequence that is generated over a number of steps. Recent top-down approaches ~\cite{khan2024cad,ma2024draw} generate a CAD sequence autoregressively, meaning that inaccurately predicted sequence tokens are fed back into the model, potentially compounding errors and degrading the accuracy of subsequent token predictions. Similarly, bottom-up approaches ~\cite{uy2022point2cyl,li2023secad,ren2022extrudenet} involve multiple sequential stages, such as per-point segmentation, base or barrel label prediction, and extrusion parameter estimation, with error propagating at each stage. Nevertheless, to assess the robustness of the different steps of \modelname, we evaluate its core stages independently (plane detection, sketch parameterization, and extrusion parameter optimization) in Table~\ref{tab:error_analysis}. In each case, we observe improved performance compared to the corresponding stages in related methods, which explains the superior overall performance achieved by our approach.

\textbf{Plane Detection.} We compare \modelname’s plane detection performance against CAD-SIGNet [15], using standard precision, recall, and F1-score. Note that CADSIG-Net is designed to align plane detection with design history, hence its predicted and ground-truth planes are anchored in the original DeepCAD design planes. In contrast, \modelname~targets reverse engineering and thus operates on cross-sectional planes obtained by pre-processing DeepCAD (detailed in Supp. Section~\ref{sec:data_preprocess}). The Table~\ref{tab:error_analysis} shows that \modelname~achieves strong performance in cross-sectional plane detection, with notably higher metrics than CAD-SIGNet’s performance in the design-plane detection setting.

\textbf{Sketch Parameterization.} We extend our evaluation on Table 5 of the main paper with the addition of an autoregressive baseline, Vitruvion [4]. We show in the Table~\ref{tab:error_analysis} that \modelname~outperforms both Vitruvion [4] and Davinci [35] in cross-section parameterization in terms of Sketch Chamfer Distance (SCD).

\textbf{Extrusion Performance.} To evaluate extrusion performance, we compare \modelname with Point2Cyl~\cite{uy2022point2cyl}, a bottom-up method that estimates extrusion cylinders from 3D point clouds. We use 100 CAD models from the dataset provided in [8] and, for each model, match each ground-truth extrusion cylinder to its closest predicted counterpart from both methods using Chamfer Distance (CD). We then report the average CD across all matched pairs. As shown in the Table~\ref{tab:error_analysis}, \modelname~achieves a lower average CD, indicating more accurate extrusion recovery. It is worth noting that, due to the high computational cost of Point2Cyl (over 4 hours for 100 samples), this evaluation was limited to 100 models.

\begin{table*}[h]
\centering
\resizebox{\textwidth}{!}{%
\begin{tabular}{lccc}
\toprule
\multicolumn{4}{c}{\textbf{Plane Detection}} \\
\midrule
Method & Precision & Recall & F1 \\
\midrule
CADSIG-Net~\cite{khan2024cad} & 0.701 & 0.696 & 0.686 \\
Ours                 &\textbf{ 0.894} & \textbf{0.864} &\textbf{ 0.870} \\
\bottomrule
\end{tabular}
\hspace{1cm}
\begin{tabular}{lc}
\toprule
\multicolumn{2}{c}{\textbf{Extrusion Performance}} \\
\midrule
Method & Extrusion CD\textdownarrow \\
\midrule
Point2Cyl~\cite{uy2022point2cyl} & 27.9 \\
Ours               & \textbf{10.1} \\
\bottomrule
\end{tabular}
\hspace{1cm}
\vspace{1cm}
\begin{tabular}{lc}
\toprule
\multicolumn{2}{c}{\textbf{Sketch Parameterization}} \\
\midrule
Method & SCD\textdownarrow \\
\midrule
Vitruvion~\cite{seff2022vitruvion} & 1.236 \\
Davinci~\cite{karadeniz2024davinci}  & 0.827 \\
Ours               & \textbf{0.283} \\
\bottomrule
\end{tabular}
}
\caption{Error accumulation analysis across plane detection, sketch parameterization, and extrusion performance.}
\label{tab:error_analysis}
\end{table*}

\subsubsection{Cut Extrusions}

\begin{figure}[h]
    \centering
    \includegraphics[width=\linewidth]{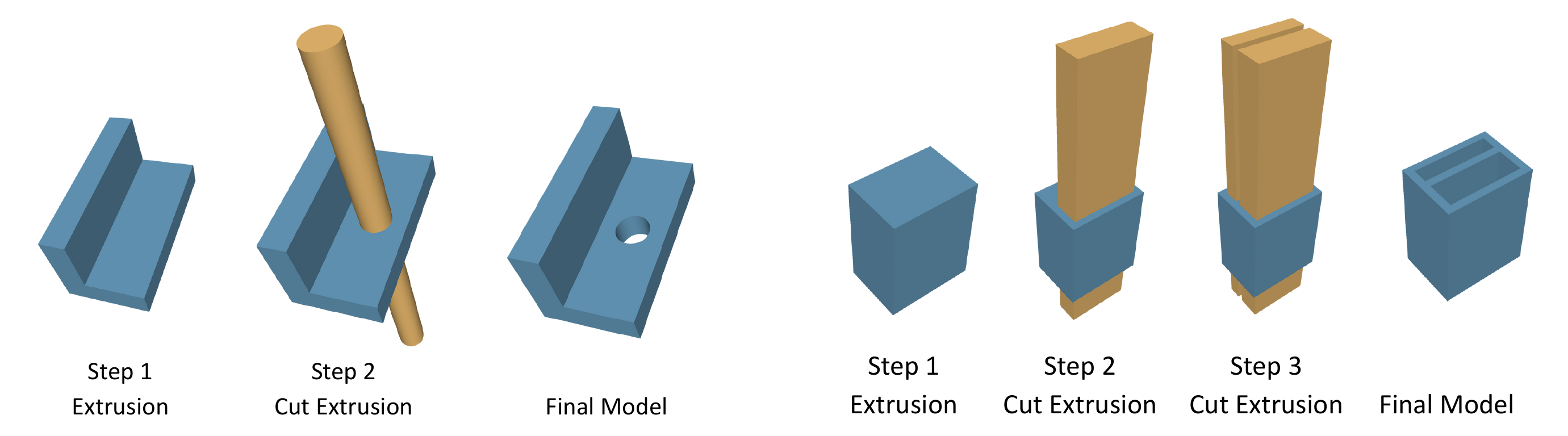}
        \vspace{-0.5cm}
    \caption{Example CAD reconstructions with cut extrusions.}
    \label{fig:cut_extrusions}
\end{figure}

Cut extrusions can be categorized into: (1) visible cut extrusions, when their originating sketch loops are visible and nested within other loops; (2) invisible cut extrusions, when their sketch loops are invisible in the final CAD geometry hence not captured by cross-sections. Our method handles visible cut extrusions as detailed in Section 4.4. The invisible cut extrusions are not directly handled. In such cases, the models are reverse engineered using alternative sketch–extrude sequences. Table~\ref{tab:cut_extrusions} demonstrates that our method outperforms CAD-SIGNet in reconstructing models with visible and invisible cut extrusions. Visual illustrations of individual reconstruction steps with cut extrusions are shown in Figure~\ref{fig:cut_extrusions}.

\begin{table}[h]
\centering
\begin{tabular}{lccccc}
\toprule
Method & Mean CD\textdownarrow & Median CD\textdownarrow & IR\textdownarrow & IoU\textuparrow & ECD\textdownarrow \\
\midrule
CAD-SIGNet~\cite{khan2024cad} & 4.95 & 0.77 & \textbf{3.0} & 65.5 & 3.97 \\
Ours                 & \textbf{4.26} & \textbf{0.64} & 3.5 & \textbf{71.8} & \textbf{3.13} \\
\bottomrule
\end{tabular}
\caption{Performance on DeepCAD models with cut extrusions.}
\label{tab:cut_extrusions}
\end{table}

\subsection{Limitations and Failure Case Analysis}
\label{sec:limitations}

The proposed method is currently limited to extrusion-based CAD operations, consistent with prior work~\cite{khan2024cad,li2023secad,uy2022point2cyl}. While extrusions are foundational to CAD modeling, many real-world designs rely on additional features such as revolutions, sweeps, lofts, and fillets. In the proposed method, extrusions are defined relative to sketch plane normals, which can lead to inaccurate outputs when the input geometry involves non-axis-aligned extrusions. Additionally, the proposed method does not support advanced sketch primitives such as B-splines, limiting its applicability to freeform designs. Future work will address these limitations by expanding the set of supported CAD operations, optimizing direction vectors in the differentiable extrusion process, and extending the sketch parameterization network to handle B-spline primitives.

\begin{figure}[h]
    \centering
    \includegraphics[width=\linewidth]{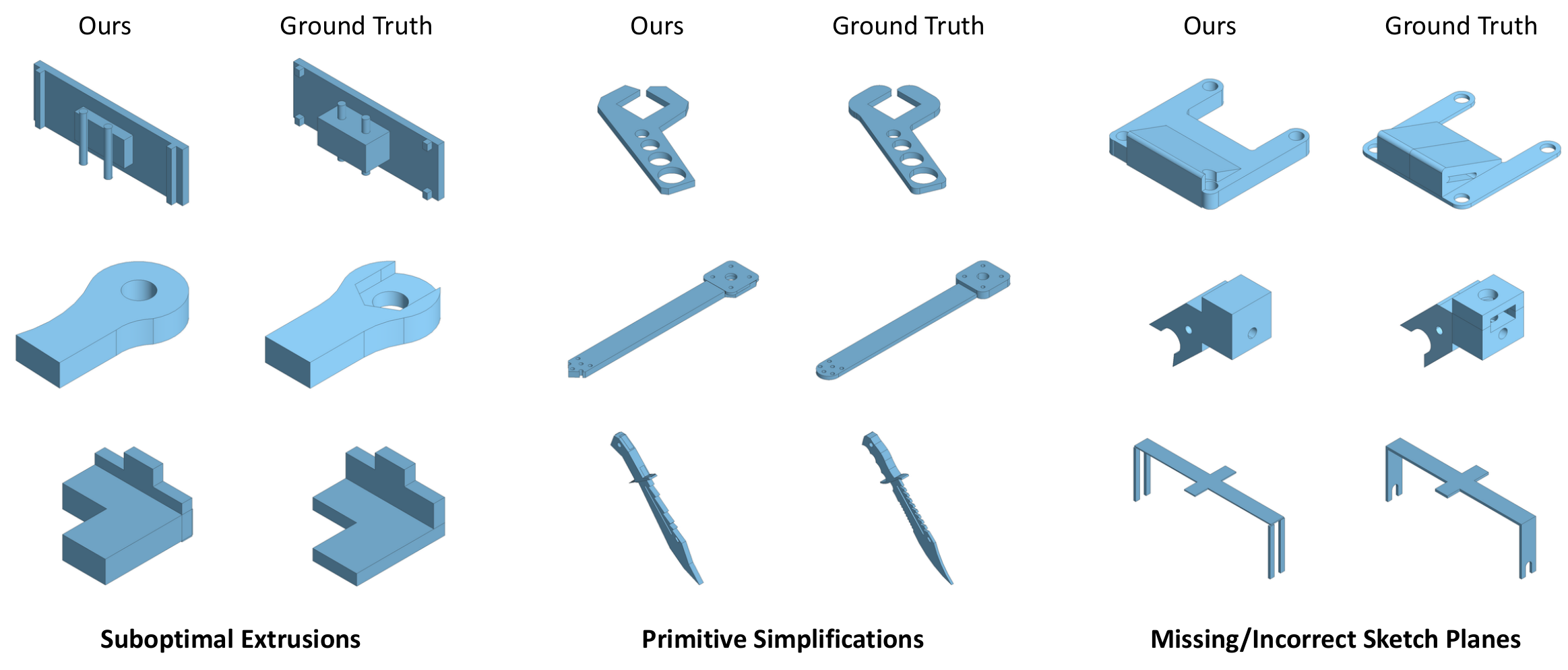}
    \caption{Representative failure cases illustrating issues such as inaccurate extrusion height, primitive simplification, and sketch plane misprediction.}

    \label{fig:failure_cases}
\end{figure}

The proposed method also exhibits some imperfections within its supported scope. We conduct a qualitative analysis of representative failure cases, as illustrated in Figure~\ref{fig:failure_cases}. Common issues include: (i) suboptimal extrusions, where the predicted extrusion height deviates from the ground truth; (ii) primitive simplifications, in which sketch elements such as arcs are approximated using coarse line segments, and (iii) missing or incorrectly predicted sketch planes, resulting in misaligned or absent features in the final 3D reconstruction. These examples highlight the difficulty of robustly predicting sketch planes, constrained sketches, and extrusion parameters, especially in models with complex geometry and topology.

\clearpage

\subsection{Additional Qualitative Results}
\label{sec:more_qualitative}

\begin{figure}[h]
    \centering
    \includegraphics[width=\linewidth]{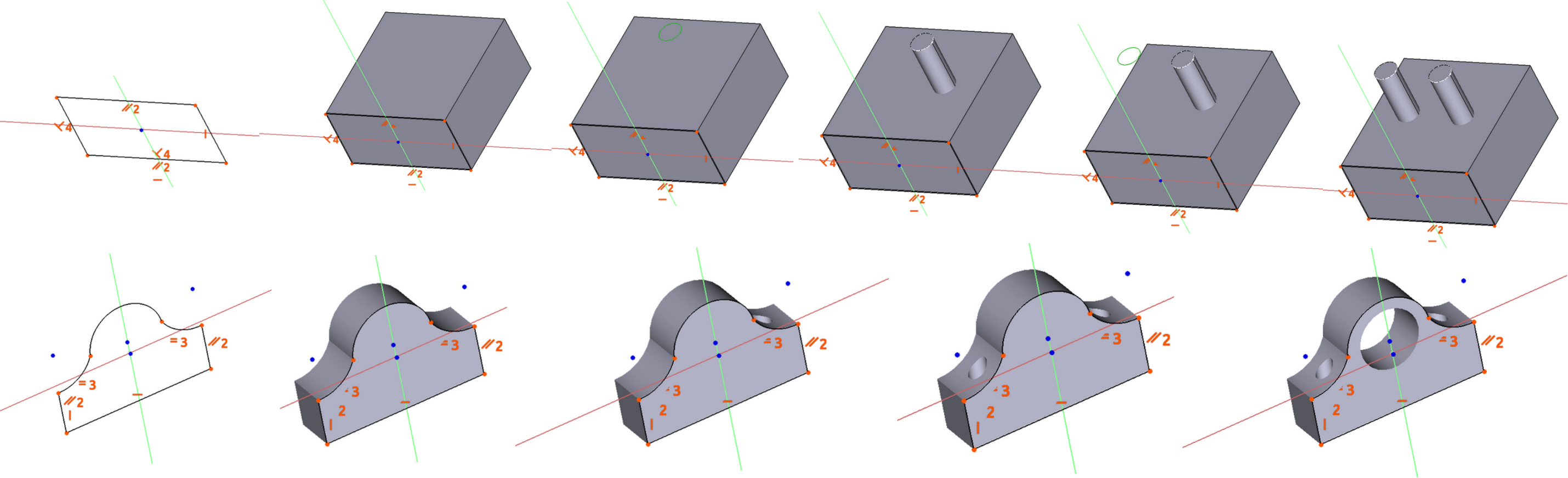}
        \vspace{-0.5cm}
    \caption{Visualization of intermediate steps in constrained CAD reconstructions in FreeCAD.}
    \label{fig:freecad_intermediate}
\end{figure}

\begin{figure}[h]
    \centering
    \includegraphics[width=\linewidth]{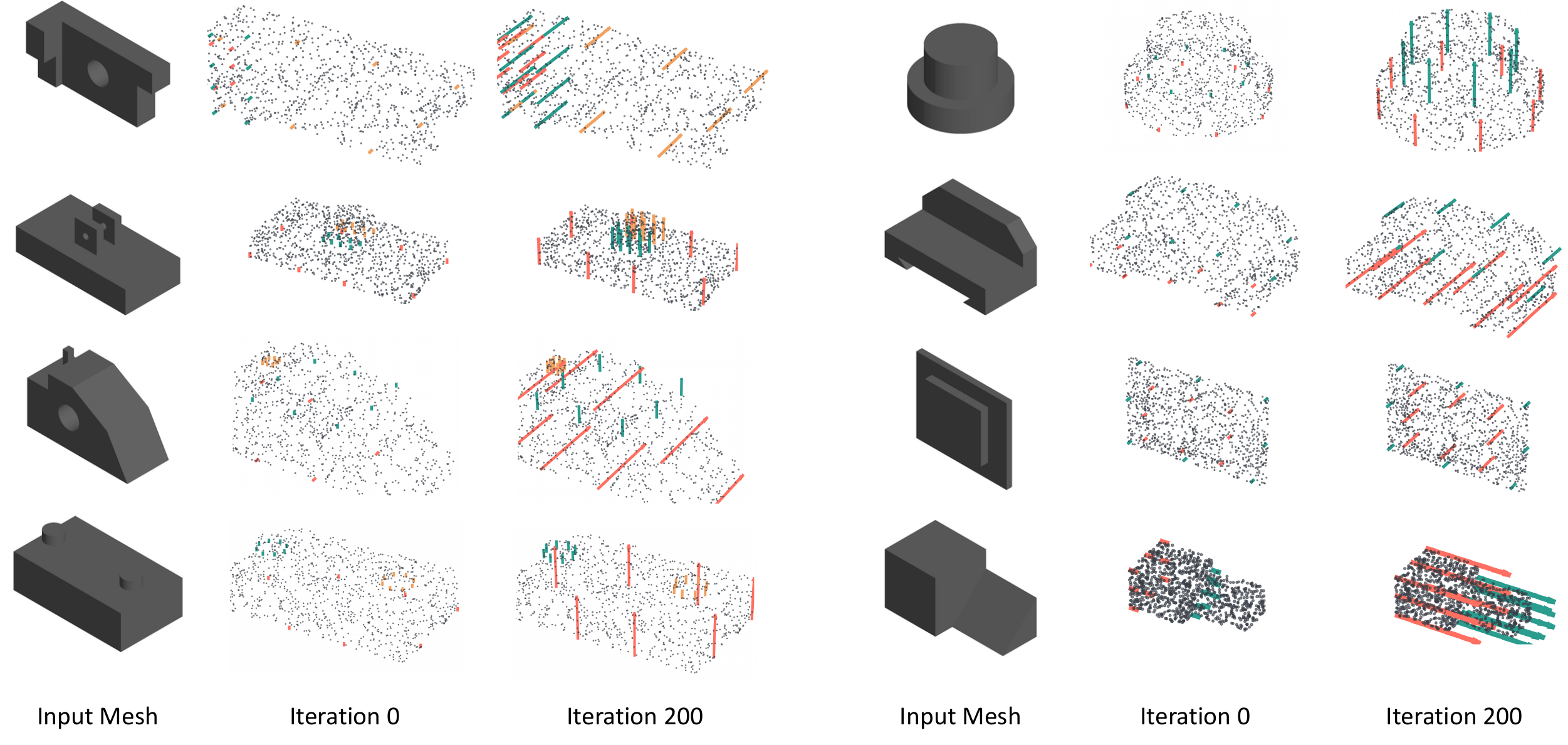}
    \vspace{-0.3cm}
    \caption{Visualizations of extrusion optimization with the points sampled from the input mesh and the extrusion vectors.}
    \label{fig:diff_optim}
\end{figure}

\begin{figure}[h]
    \centering
    \setlength{\belowcaptionskip}{-0.5cm}
    \includegraphics[width=0.95\linewidth]{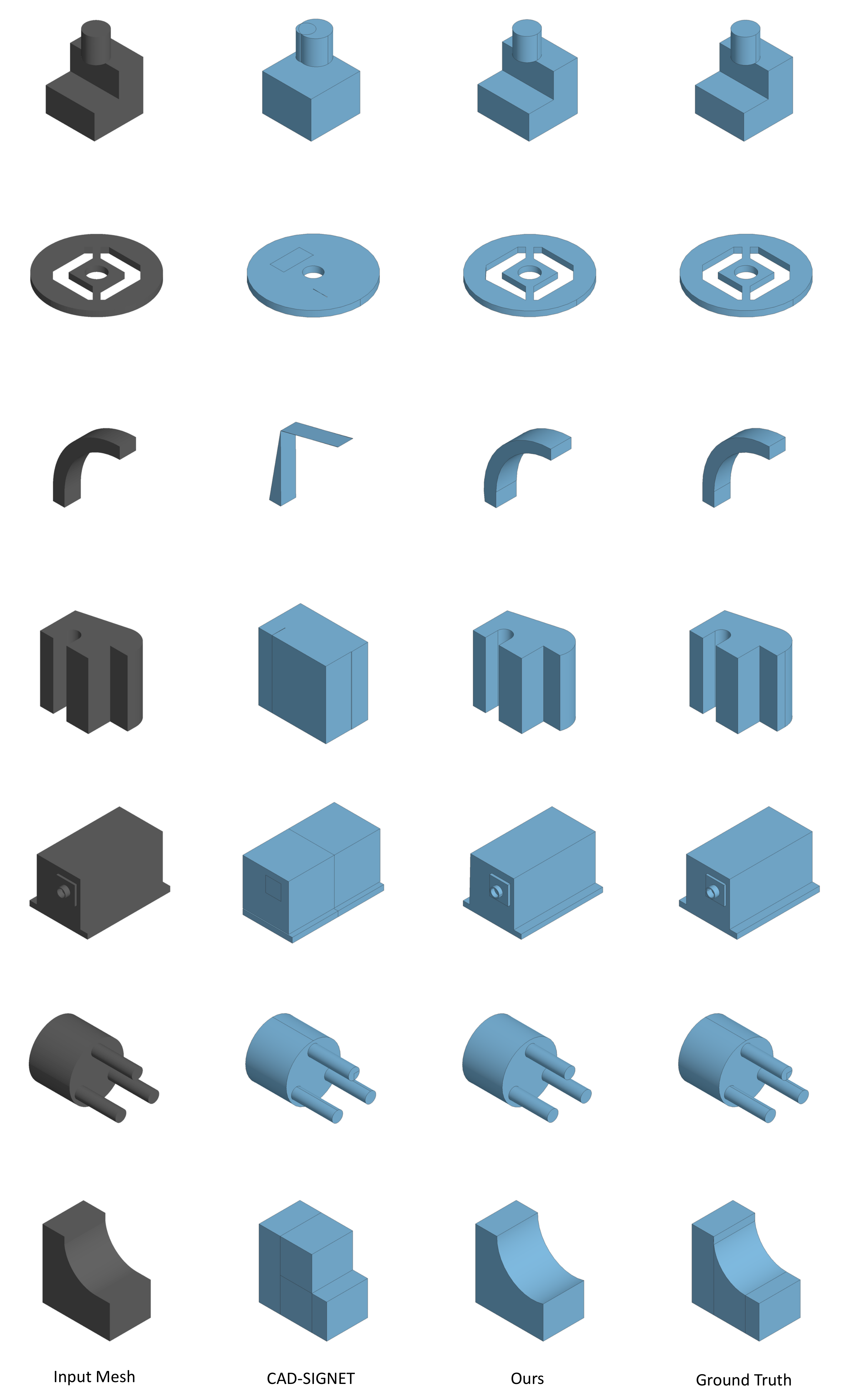}
    \caption{More qualitative comparisons with ~\cite{khan2024cad} on DeepCAD dataset.}
    \label{fig:more_qual_deepcad1}
\end{figure}

\begin{figure}[h]
    \centering
    \setlength{\belowcaptionskip}{-0.5cm}
    \includegraphics[width=\linewidth]{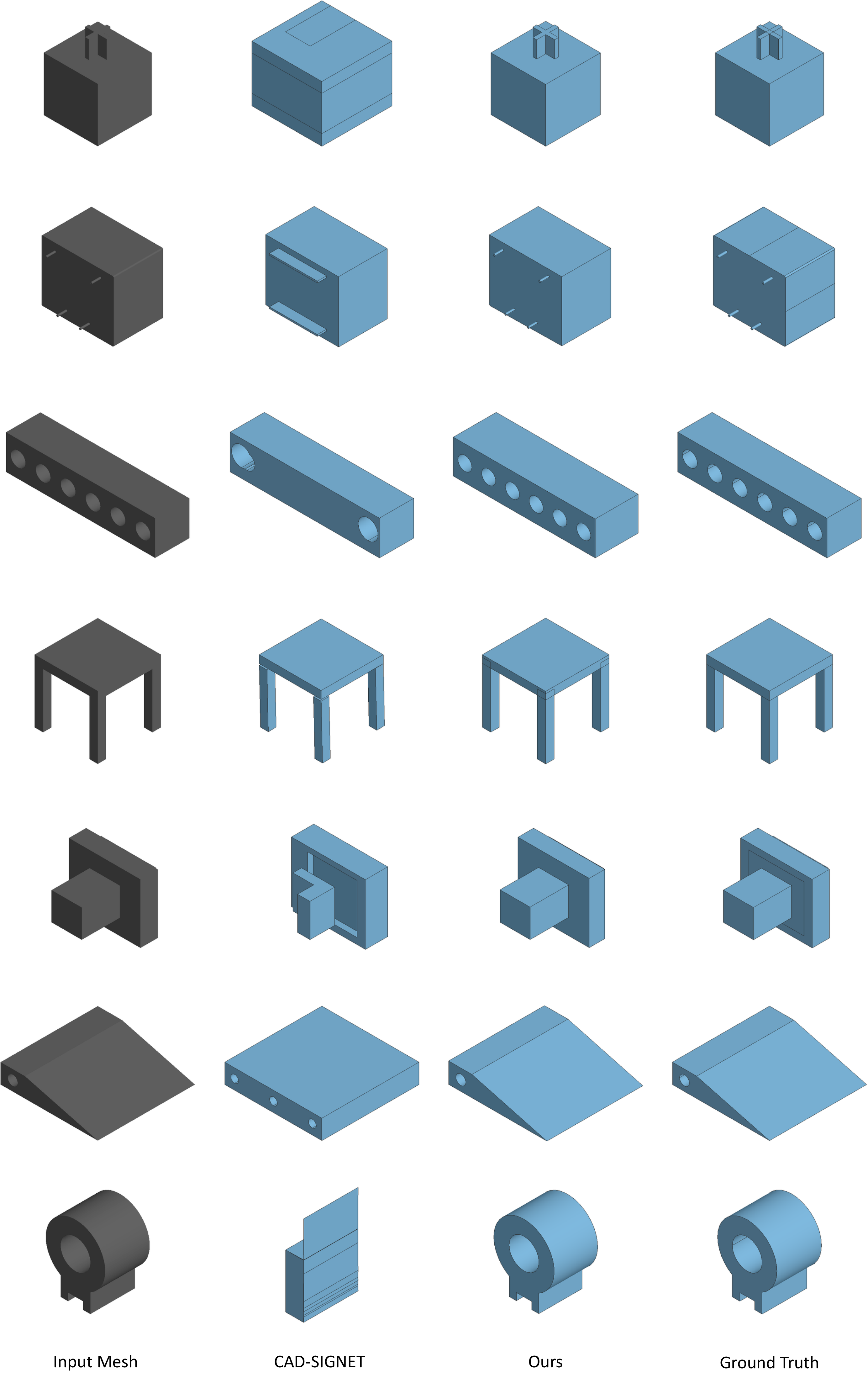}
    \caption{More qualitative comparisons with ~\cite{khan2024cad} on DeepCAD dataset.}
    \label{fig:more_qual_deepcad2}
\end{figure}

\begin{figure}[h]
    \centering
    \setlength{\belowcaptionskip}{-0.5cm}
    \includegraphics[width=\linewidth]{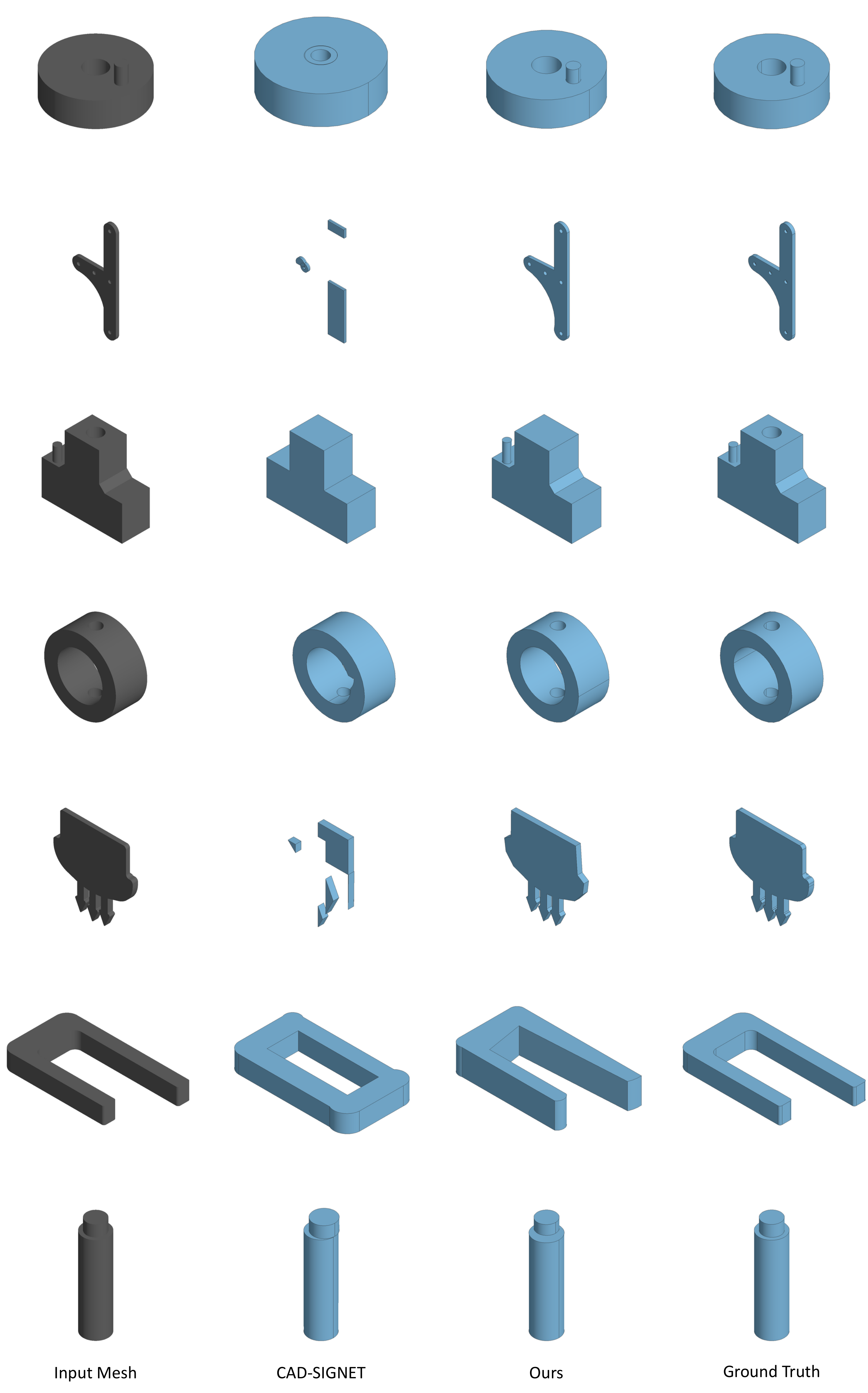}
    \caption{More qualitative comparisons with ~\cite{khan2024cad} on Fusion360 dataset.}
    \label{fig:more_qual_f360_1}
\end{figure}

\begin{figure}[h]
    \centering
    \setlength{\belowcaptionskip}{-0.5cm}
    \includegraphics[width=\linewidth]{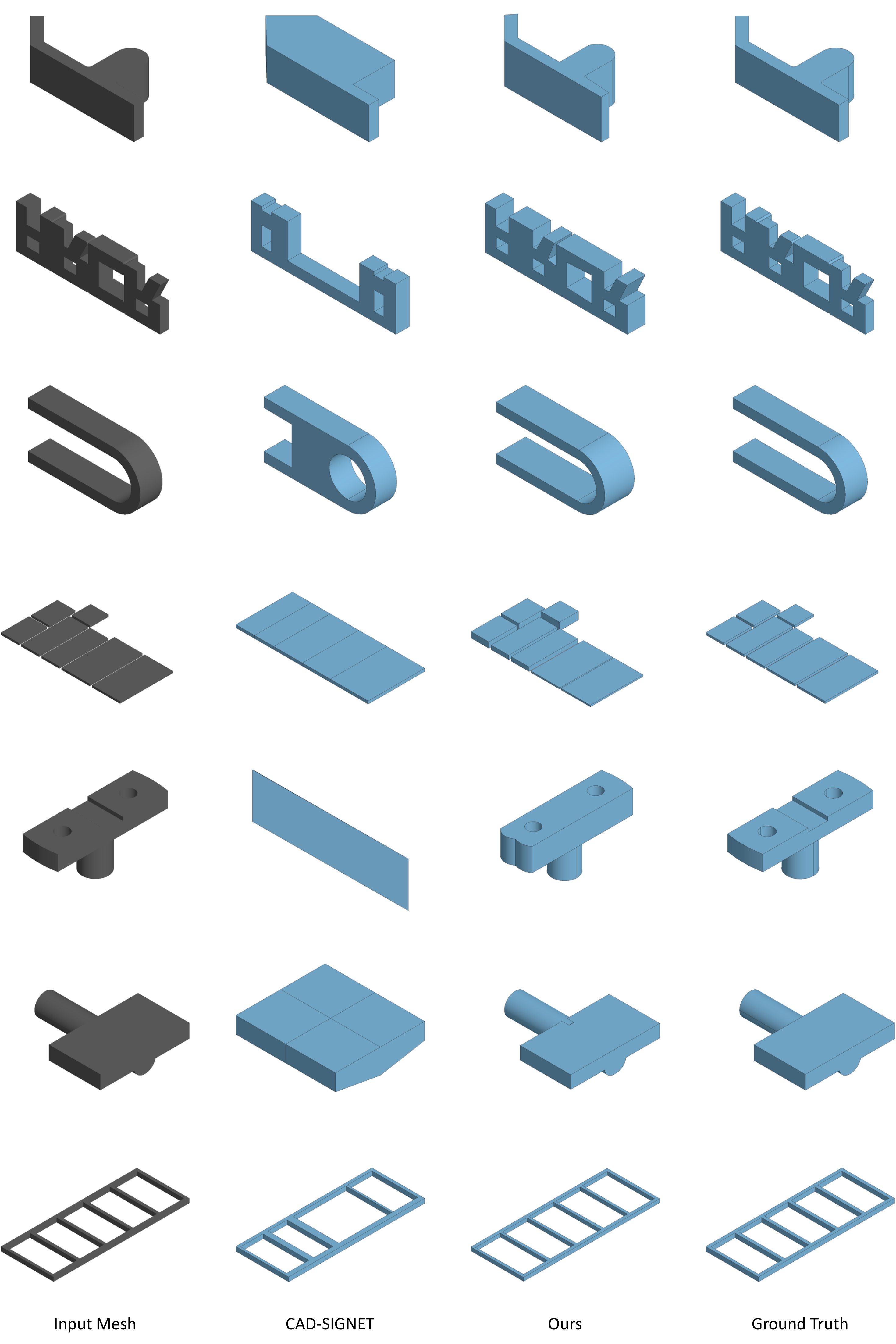}
    \caption{More qualitative comparisons with ~\cite{khan2024cad} on Fusion360 dataset.}
    \label{fig:more_qual_f360_2}
\end{figure}

\vspace{-0.3cm}
This section provides additional qualitative results and visualizations. Figure~\ref{fig:freecad_intermediate} shows intermediate steps of reconstructed CAD models in FreeCAD~\cite{FreeCAD}. Figure~\ref{fig:diff_optim} shows the visualization of differentiable extrusion. Figure~\ref{fig:more_qual_deepcad1}, Figure~\ref{fig:more_qual_deepcad2}, Figure~\ref{fig:more_qual_f360_1}, and Figure~\ref{fig:more_qual_f360_2} shows more qualitative comparisons on DeepCAD and Fusion360~\cite{willis2020fusion} datasets.

\vspace{2cm}

\vspace{2cm}

\end{document}